\documentclass{article}

\usepackage[nonatbib,final]{neurips_2018}

\usepackage[utf8]{inputenc} 
\usepackage[T1]{fontenc}    
\usepackage{hyperref}       
\usepackage{url}            
\usepackage{booktabs}       
\usepackage{amsfonts}       
\usepackage{amsmath}
\usepackage{nicefrac}       
\usepackage{microtype}      
\usepackage{color}
\usepackage{graphicx}
\usepackage{bm}
\usepackage{amsmath}
\usepackage{bbold}
\usepackage{caption}
\usepackage{subcaption}
\usepackage{wrapfig}
\usepackage[sort,numbers]{natbib}
\usepackage{multirow}
\usepackage{xspace}

\usepackage{authblk}

\title{Constrained Graph Variational Autoencoders for Molecule Design}

\author[$\enspace$1]{{\bf Qi Liu}\thanks{work performed during an internship with Microsoft Research, Cambridge.}}
\author[2]{{\bf Miltiadis Allamanis}}
\author[2]{{\bf Marc Brockschmidt}}
\author[2]{{\bf Alexander L. Gaunt}}
\affil[$\enspace$1]{Singapore University of Technology and Design}
\affil[2]{Microsoft Research, Cambridge}
\affil[ ]{\texttt{qiliu@u.nus.edu}, \texttt{\{miallama, mabrocks, algaunt\}@microsoft.com}}

\newcommand{\numatoms}{N}
\newcommand{\normal}[2]{\mathcal{N}\left(#1, #2\right)}
\newcommand{\latent}[1]{\mathbf{z}_{#1}}
\newcommand{\nodestate}[2]{\mathbf{h}^{(#1)}_#2}
\newcommand{\node}[0]{\ensuremath{v}}
\newcommand{\nodeTwo}[0]{\ensuremath{u}}
\newcommand{\nodeThree}[0]{\ensuremath{w}}
\newcommand{\nodeNetwork}{f}
\newcommand{\labelledEdge}[3]{\ensuremath{#1\mathrel{\smash{\raisebox{0pt}{\ensuremath{\stackrel{#3}{\smash{\raisebox{-1.1pt}{$\leftrightarrow$}}}}}}}#2}}
\newcommand{\edge}[2]{#1\mathrel{\leftrightarrow}#2}

\newcommand{\figref}[1]{Fig. \ref{fig:#1}}

\newcommand{\secref}[1]{Sect. \ref{sec:#1}}
\newcommand{\eqnref}[1]{Eq. \ref{eqn:#1}}
\newcommand{\ourmodel}{CGVAE\xspace}
\newcommand{\focus}{\texttt{focus}\xspace}
\newcommand{\expand}{\texttt{expand}\xspace}
\newcommand{\yujia}{DeepGAR\xspace}

\begin{document}

\maketitle
\vspace{-0.5cm}
\begin{abstract}
  Graphs are ubiquitous data structures for representing interactions between entities.
	With an emphasis on applications in chemistry, we explore
	the task of learning to generate graphs that conform to a distribution
	observed in training data. We propose a variational autoencoder
        model in which both encoder and decoder are graph-structured.
	Our decoder assumes a sequential ordering of graph extension steps and we
  discuss and analyze design choices that mitigate the potential downsides of
  this linearization.
	Experiments compare our approach with a wide range of baselines on the molecule
	generation task and show that our method is
  successful at matching the statistics of the original dataset on semantically
  important metrics. Furthermore, we show that by using appropriate shaping of
  the latent space, our model allows us to design molecules that are (locally)
  optimal in desired properties.
		
%
\end{abstract}

\section{Introduction}
Structured objects such as program source code, physical systems, chemical molecules and even 3D scenes
are often well represented using graphs~\citep{allamanis18learning, Gilmer17,
  qi20173d, kipf2018neural}. Recently,
considerable progress has been made on building \emph{discriminative} deep
learning models that ingest graphs as inputs~\citep{gori2005new, li2015gated, defferrard2016convolutional, kipf2016semi}.
Deep learning approaches have also been
suggested for graph \emph{generation}. More specifically, generating
and optimizing chemical molecules has been identified as an important
real-world application for this set of techniques~\citep{gomez-bombarelli16automatic, olivecrona2017molecular, segler2017generating, neil2018exploring, samanta2018designing}.

In this paper, we propose a novel probabilistic model for graph generation that builds
gated graph neural networks (GGNNs)~\citep{li2015gated} into the encoder
and decoder of a variational autoencoder (VAE)~\citep{kingma2013auto}.
Furthermore, we demonstrate
how to incorporate hard domain-specific constraints into our
model to adapt it for the molecule generation task.
With these constraints in place, we refer to our model as a constrained graph variational
autoencoder (\ourmodel). Additionally, we shape the latent space of the VAE to allow optimization of
numerical properties of the resulting molecules.
Our experiments are performed with real-world datasets of molecules with
pharmaceutical and photo-voltaic applications.
By generating novel molecules from these datasets, we demonstrate the benefits of
our architectural choices. In particular, we observe that (1) the GGNN architecture is beneficial
for state-of-the-art generation of molecules matching chemically relevant statistics of the
training distribution, and (2) the semantically meaningful latent space 
arising from the VAE allows continuous optimization of
molecule properties~\citep{gomez-bombarelli16automatic}.

The key challenge in generating graphs is that sampling directly from a joint distribution
over all configurations of labeled nodes and edges is intractable for reasonably sized graphs.
Therefore, a generative model must decompose this joint in some way. 
A straightforward approximation is to ignore correlations and model the
existence and label of each edge with
\emph{independent} random variables~\citep{erdos1959random, snijders1997estimation, simonovsky2018towards}.
An alternative approach is to factor the distribution into a \emph{sequence} of
discrete decisions in a graph construction trace~\citep{li2018learning, you18}.
Since correlations between
edges are usually crucial in real applications,
we pick the latter, sequential, approach in this work.
Note that for molecule design, some correlations take the form of
known hard rules governing molecule stability, and we explicitly enforce 
these rules wherever possible using a
technique that masks out choices leading to illegal graphs~\cite{kusner2017grammar, samanta2018designing}.
The remaining ``soft'' correlations (e.g. disfavoring of small cycles) are learned by our graph structured VAE.

By opting to generate graphs sequentially, we lose permutation symmetry and have to train using
arbitrary graph linearizations. For computational reasons, we cannot 
consider all possible linearizations for each graph, so it is challenging to marginalize out
the construction trace when computing the log-likelihood of a graph in the VAE objective. We design
a generative model where the learned component is conditioned only on the current state of generation
and not on the arbitrarily chosen path to that state. We argue that this property is intuitively desirable and 
show how to derive a bound for the desired log-likelihood under this model. 
Furthermore, this property makes the model
relatively shallow and it is easy scale and train.



\section{Related Work}
Generating graphs has a long history in research, and we consider three classes of related work: Works that ignore correlations between edges, works that generate graphs sequentially and works that emphasize the application to molecule design.

\paragraph{Uncorrelated generation}
The Erd\H{o}s-R\'{e}nyi $G(n,p)$ random graph model~\citep{erdos1959random} is the simplest example of this class of algorithms, where each edge exists with independent probability $p$. Stochastic block models~\citep{snijders1997estimation} add community structure to the Erd\H{o}s-R\'{e}nyi model, but retain uncorrelated edge sampling. Other traditional random graph models such as those of \citet{albert2002statistical, leskovec2010kronecker} do account for edge correlations, but they are hand-crafted into the models. A more modern \emph{learned} approach in this class is GraphVAEs~\citep{simonovsky2018towards}, where the decoder emits independent probabilities governing edge and node existence and labels.

\paragraph{Sequential generation}
\citet{johnson2016learning} sidesteps the issue of permutation symmetry by considering the task of generating a graph from an auxiliary stream of information that imposes an order on construction steps. This work outlined many ingredients for the general sequential graph generation task: using GGNNs to embed the current state of generation and multi-layer perceptrons (MLPs) to drive decisions based on this embedding. \citet{li2018learning} uses these ingredients to build an autoregressive model for graphs without the auxiliary stream. Their model gives good results, but each decision is conditioned on a full history of the generation sequence, and the authors remark on stability and scalability problems arising from the optimization of very deep neural networks. In addition, they describe some evidence for overfitting to the chosen linearization scheme due to the strong history dependence. Our approach also uses the ingredients from \citet{johnson2016learning}, but avoids the training and overfitting problems using a model that is conditioned only on the current partial graph rather than on full generation traces. In addition, we combine Johnson's ingredients with a VAE that produces a meaningful latent space to enable continuous graph optimization \cite{gomez-bombarelli16automatic}. 

An alternative sequential generation algorithm based on RNNs is presented in~\citet{you18}. The authors point out that a dense implementation of a GGNN requires a large number $\mathcal{O}(e\numatoms^2)$ of operations to construct a graph with $e$ edges and $\numatoms$ nodes. We note that this scaling problem can be mitigated using a sparse GGNN implementation~\citep{allamanis18learning}, which reduces complexity to $\mathcal{O}(e^2)$.

\paragraph{Molecule design}
Traditional \textit{in silico} molecule design approaches rely on considerable domain knowledge, physical simulation and heuristic search algorithms (for a recent example, see \citet{gomez2016design}). Several deep learning approaches have also been tailored to molecule design, for example~\cite{jin2018junction} is a very promising method that uses a library of frequent (ring-containing) fragments to reduce the graph generation process to a tree generation process where nodes represent entire fragments. Alternatively, many methods rely on the SMILES linearization of molecules~\citep{weininger1988smiles} and use RNNs to generate new SMILES strings~\citep{olivecrona2017molecular, segler2017generating, neil2018exploring, gomez-bombarelli16automatic}. A particular challenge of this approach is to ensure that the generated strings are syntactically valid under the SMILES grammar. The Grammar VAE uses a mask to impose these constraints during generation and a similar technique is applied for general graph construction in~\citet{samanta2018designing}. Our model also employs masking that, among other things, ensures that the molecules we generate can be converted to syntactically valid SMILES strings.

\section{Generative Model}
\label{sec:GenProc}
\begin{figure}
\centering
\includegraphics[width=0.85\textwidth,clip,trim=0cm 5cm 03cm 1cm]{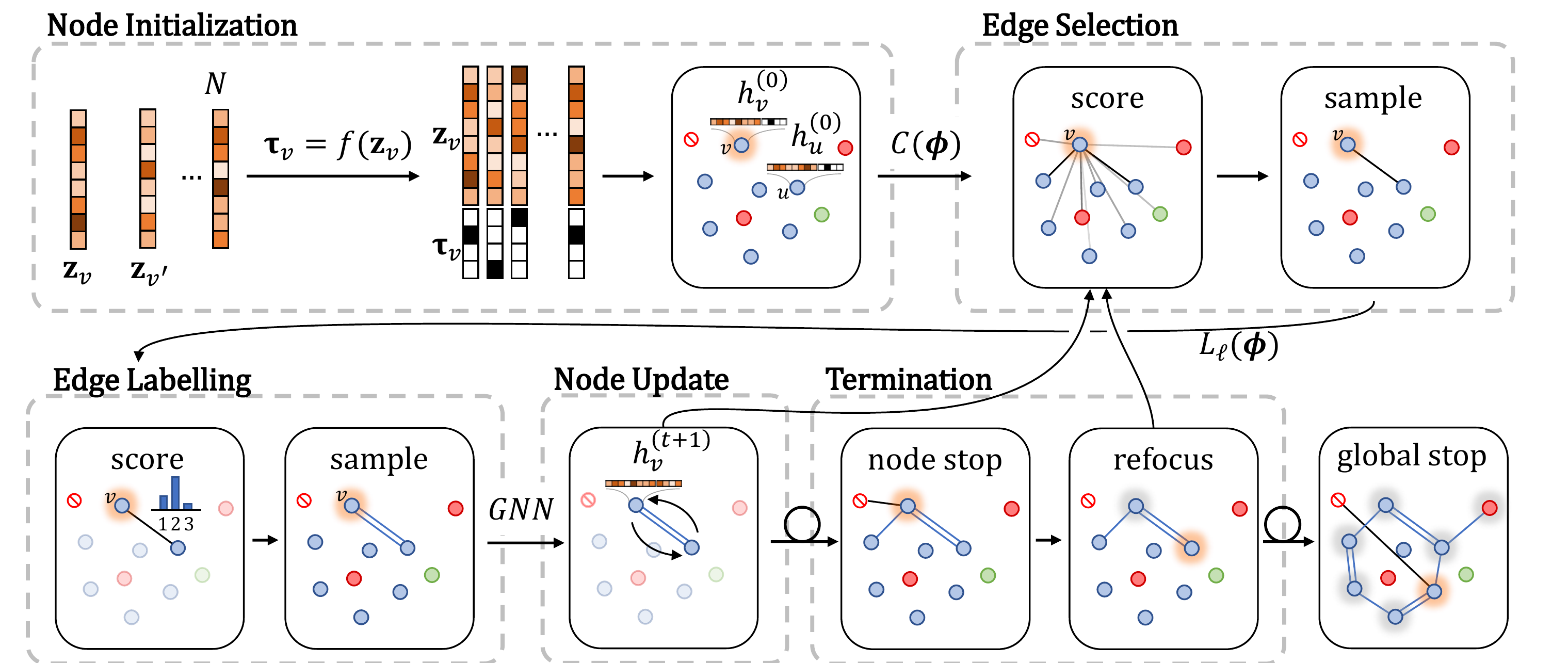}
\caption{\label{fig:decoder}Illustration of the phases of the generative
  procedure. Nodes are initialized with latent variables and then we enter a
  loop between edge selection, edge labelling and node
  update steps until the special stop node $\oslash$ is selected. We then refocus to
  a new node or terminate if there are no candidate focus nodes in the
  connected component. A looped arrow indicates that several loop iterations may
  happen between the illustrated steps.}
\end{figure}
Our generative procedure is illustrated in \figref{decoder}. The process is seeded with $\numatoms$ vectors $\latent{\node}$ that together form a
latent ``specification'' for the graph to be generated ($\numatoms$ is an upper bound on the number of nodes in the final graph).
Generation of edges between these nodes then proceeds using two decision functions: \focus and \expand. In each step the \focus function chooses a focus node to visit, and then the \expand function chooses edges to add from the focus node. As in breadth-first traversal, we implement \focus as a deterministic queue (with a random choice for the initial node).

Our task is thus reduced to learning the \expand function that enumerates new
edges connected to the currently focused node.
One design choice is to make \expand condition upon
the full history of the generation.
However, this has both theoretical and practical downsides.
Theoretically, this means that the learned model is likely to learn to reproduce
generation traces.
This is undesirable, since the underlying data usually only contains fully formed
graphs; thus the exact form of the trace is an artifact of the implemented
data preprocessing.
Practically, this would lead to extremely deep computation graphs, as even small
graphs easily have many dozens of edges; this makes training of the resulting
models very hard as mentioned in mentioned in \citet{li2018learning}.
Hence, we condition \expand only upon the partial graph structure
$\smash{\mathcal{G}^{(t)}}$ generated so far; intuitively, this corresponds to
learning how to complete a partial graph without using any information
about how the partial graph was generated.
We now present the details of each stage of this generative procedure.

\paragraph{Node Initialization}
We associate a state $\smash{\nodestate{t=0}{\node}}$ with each node $\node$
in a set of initially unconnected nodes. 
Specifically, $\latent{\node}$ is drawn from the $d$-dimensional standard
normal $\normal{\mathbf{0}}{\mathbf{I}}$, and $\smash{\nodestate{t=0}{\node}}$
is the concatenation $[\latent{\node}, \bm{\tau}_\node]$, where
$\bm{\tau}_\node$ is an interpretable one-hot vector indicating the node
type. $\bm{\tau}_\node$ is derived from $\latent{\node}$ by sampling from the
softmax output of a learned mapping $\bm{\tau}_\node \sim
\nodeNetwork(\latent{\node})$ where $\nodeNetwork$ is a neural
network\footnote{We implement $f$ as a linear classifier from the $100$ dimensional latent space to one of the node type classes.}.
The interpretable component of $\smash{\nodestate{t=0}{\node}}$ gives us a means to
enforce hard constraints during generation.

From these node-level variables, we can calculate global
representations $\smash{\mathbf{H}^{(t)}}$ (the average representation
of nodes in the connected component at generation step $t$), and
$\mathbf{H}_{\rm init}$ (the average representation of all nodes at $t=0$).
In addition to $\numatoms$ working nodes, we also initialize a special ``stop
node'' to a learned representation $\mathbf{h}_{\oslash}$ for managing algorithm
termination (see below).

\paragraph{Node Update}
Whenever we obtain a new graph $\mathcal{G}^{(t+1)}$, we discard
$\smash{\nodestate{t}{\node}}$ and compute new representations
$\smash{\nodestate{t+1}{\node}}$ for all nodes taking their (possibly changed)
neighborhood into account.
This is implemented using a standard gated graph neural network (GGNN) $G_{\rm
  dec}$ for $S$ steps\footnote{Our experiments use $S=7$.}, which is defined as a recurrent operation over messages
$\smash{\mathbf{m}_\node^{(s)}}$.
\begin{align*}
  \mathbf{m}^{(0)}_\node = \nodestate{0}{\node}
  &&
  \mathbf{m}^{(s+1)}_\node =
     {\rm GRU}\left[\mathbf{m}^{(s)}_\node,
               \sum_{\labelledEdge{\node}{\nodeTwo}{\ell}} E_\ell(\mathbf{m}^{(s)}_\nodeTwo)\right]
  &&
     \nodestate{t+1}{\node} = \mathbf{m}^{(S)}_\node
\end{align*}
Here the sum runs over all edges in the current graph and $E_\ell$ is an edge-type
specific neural network\footnote{In our implementation, $E_\ell$ is a dimension-preserving linear transformation.}
We also augment our model with a master node as described by \citet{Gilmer17}.
Note that since $\smash{\nodestate{t+1}{\node}}$ is computed from $\smash{\nodestate{0}{\node}}$
rather than $\smash{\nodestate{t}{\node}}$, the representation $\smash{\nodestate{t+1}{\node}}$ is independent of
the generation history of $\mathcal{G}^{(t+1)}$.

\paragraph{Edge Selection and Labelling}
We first pick a focus node $\node$ from our queue. The function \expand
then selects edges $\labelledEdge{\node}{\nodeTwo}{\ell}$
from $\node$ to $\nodeTwo$ with label $\ell$ as follows.
For each non-focus node $\nodeTwo$, we construct a feature
vector
 $\smash{\bm{\phi}_{\node,\nodeTwo}^{(t)} = [\mathbf{h}^{(t)}_\node, \mathbf{h}^{(t)}_\nodeTwo, d_{\node, \nodeTwo}, \mathbf{H}_{\rm init}, \mathbf{H}^{(t)}]}$, where $d_{\node, \nodeTwo}$ is the graph distance between $\node$ and $\nodeTwo$.
This provides the model with both
 local information for the focus node $v$ and the candidate edge
  ($\smash{\mathbf{h}_{\node}^{(t)}}, \smash{\mathbf{h}_{\nodeTwo}^{(t)}}$), and 
 global information regarding the original graph specification
  ($\smash{\mathbf{H}_{\rm init}}$) and the current graph state
  ($\smash{\mathbf{H}^{(t)}}$).
We use these representations to produce a distribution over candidate edges:
\begin{align*}
  p(\labelledEdge{\node}{\nodeTwo}{\ell} \mid \smash{\bm{\phi}_{\node,\nodeTwo}	^{(t)}})
  =
   p(\ell \mid \smash{\bm{\phi}_{\node,\nodeTwo}^{(t)}}, \edge{\node}{\nodeTwo})
   \cdot   
   p(\edge{\node}{\nodeTwo} \mid \smash{\bm{\phi}_{\node,\nodeTwo}^{(t)}}).
\end{align*}
The factors are calculated as softmax outputs from neural networks $C$
(determining the target node for an edge) and $L_\ell$ (determining the type of
the edge):\footnote{$C$ and $L_\ell$ are fully connected networks with a single hidden layer of 200 units and ReLU non-linearities.}
\begin{align}
  p(\edge{\node}{\nodeTwo} \mid \smash{\bm{\phi}_{\node,\nodeTwo}^{(t)}}) =
   \frac{M^{(t)}_{\edge{\node}{\nodeTwo}}\exp[C(\bm{\phi}_{\node,\nodeTwo}^{(t)})]}
        {\sum_{\nodeThree} M^{(t)}_{\edge{\node}{\nodeThree}}\exp[C(\bm{\phi}_{\node,\nodeThree}^{(t)})]},
&&
  p(\ell \mid \bm{\phi}_{\node,\nodeTwo}^{(t)}) =
   \frac{m^{(t)}_{\labelledEdge{\node}{\nodeTwo}{\ell}}\exp[L_\ell(\bm{\phi}_{\node,\nodeTwo}^{(t)})]}
        {\sum_{k} m^{(t)}_{\labelledEdge{\node}{\nodeTwo}{k}}\exp[L_{k}(\bm{\phi}_{\node,\nodeTwo}^{(t)})]}.
\label{eqn:factoredProb}
\end{align}
$M^{(t)}_{\edge{\node}{\nodeTwo}}$ and $m^{(t)}_{\labelledEdge{\node}{\nodeTwo}{\ell}}$
are binary masks that forbid edges that violate constraints. We discuss the construction of these masks
for the molecule generation case in \secref{masking}.
New edges are sampled from these distributions, and any nodes that are
connected to the graph for the first time are added to the \focus queue.
Note that we only consider undirected edges in this paper, but it is easy to extend the model to directed graphs.

\paragraph{Termination}
We keep adding edges to a node $\node$ using \expand and $G_{\rm
  dec}$ until an edge to the stop node is selected. Node $\node$ then loses focus and becomes ``closed'' (mask $M$ ensures that no further
edges will ever be made to $\node$). The next focus node is selected from the \focus queue.
In this way, a single connected component is grown in a breadth-first manner. Edge generation
continues until the queue is empty (note that this may leave some
unconnected nodes that will be discarded).

\section{Training the Generative Model}
The model from \secref{GenProc} relies on a latent space with
semantically meaningful points concentrated in the region weighted under the
standard normal, and trained networks $\nodeNetwork$, $C$, $L_\ell$ and $G_{\rm
  dec}$. We train these in a VAE architecture on a large dataset $\mathcal{D}$ of graphs. 
Details of this VAE are provided below.
  

\subsection{Encoder} The encoder of our VAE is a GGNN $G_{\rm enc}$ that embeds each node in an input graph $\mathcal{G}$ to a diagonal normal distribution in $d$-dimensional latent space parametrized by mean $\bm{\mu}_\node$ and standard deviation $\bm{\sigma}_\node$ vectors. The latent vectors $\latent{\node}$ are sampled from these distributions, and we construct the usual VAE regularizer term measuring the KL divergence between the encoder distribution and the standard Gaussian prior: $
\mathcal{L}_{\rm latent} = \sum_{\node\in \mathcal{G}} {\rm KL}(\normal{\bm{\mu}_\node}{{\rm diag}(\bm{\sigma}_\node)^2} ||\; \normal{\mathbf{0}}{\mathbf{I}})$.

\subsection{Decoder}
\label{sec:Decoder}
The decoder is the generative procedure described in \secref{GenProc}, and we 
condition generation on a latent sample from the encoder distribution during training.
We supervise training of the overall model using generation traces extracted from graphs in
$\mathcal{D}$.

\paragraph{Node Initialization}
To obtain initial node states $\smash{\nodestate{t=0}{\node}}$, we first sample
a node specification $\latent{\node}$ for each node $\node$ and then 
independently for each node we generate
the label $\bm{\tau}_\node$ using the learned function $f$.
The probability of re-generating the labels $\bm{\tau}^*_v$ observed in the
encoded graph is given by a sum over node permutations $\mathcal{P}$:
\begin{equation*}
p(\mathcal{G}^{(0)} \mid \latent{})=
\sum_{\mathcal{P}}
  p(\bm{\tau} = \mathcal{P}(\bm{\tau^*}) \mid \latent{})>\prod_\node p(\bm{\tau}_\node=\bm{\tau}^*_\node \mid \latent{\node}).
\end{equation*}
This inequality provides a lower bound given by the single contribution from the
ordering used in the encoder (recall that in the encoder we know the node type
$\bm{\tau}^*_\node$ from which $\latent{\node}$ was generated).
A set2set model \cite{vinyals2015order} could improve this bound.

\paragraph{Edge Selection and Labelling}
During training, we provide supervision on the sequence of edge additions based
on breadth-first traversals of each graph in the dataset $\mathcal{D}$.
Formally, to learn a distribution over \emph{graphs} (and not graph generation
traces), we would need to train with an objective that computes the
log-likelihood of each graph by marginalizing over all possible breadth-first
traces.
This is computationally intractable, so in practice we only compute a
Monte-Carlo estimate of the marginal on a small set of sampled traces.
However, recall from \secref{GenProc} that our \expand model is not conditioned
on full traces, and instead only considers the partial graph generated so far.
Below we outline how this intuitive design formally affects the VAE training
objective.

Given the initial collection of unconnected nodes, $\mathcal{G}^{(0)}$, from the initialization
above, we first use Jensen's inequality to show that the log-likelihood of a
graph $\mathcal{G}$ is loosely lower bounded by the expected log-likelihood of
all the traces $\Pi$ that generate it.
\begin{equation}
\log p(\mathcal{G}  \mid  \mathcal{G}^{(0)}) 
=
\log \sum_{\pi\in\Pi}p(\pi  \mid  \mathcal{G}^{(0)}) 
\geq
\log(|\Pi|) + \frac{1}{|\Pi|} \sum_{\pi\in\Pi} \log p(\pi  \mid  \mathcal{G}^{(0)})
\label{eqn:graphLogProb}
\end{equation}
We can decompose each full generation trace $\pi \in \Pi$ into a sequence of
steps of the form $(t, \node, \epsilon)$, where $\node$ is the current focus
node and $\epsilon = \labelledEdge{\node}{\nodeTwo}{\ell}$ is the edge added at
step $t$:
\begin{equation*}
\log p(\pi \mid \mathcal{G}^{(0)})
=\sum_{(t, \node,\epsilon)\in \pi} 
	  \left\{
		\log p(\node \mid \pi, t)
		+
		\log p(\epsilon \mid \mathcal{G}^{(t-1)}, \node)
		\right\}
\end{equation*}
The first term corresponds to the choice of $v$ as focus node at step $t$ of
trace $\pi$. As our \focus{} function is fixed, this choice is uniform in the first
focus node and then deterministically follows a breadth-first queuing system.
A summation over this term thus evaluates to the constant $\log(1/\numatoms)$.

As discussed above, the second term is only conditioned on the current graph
(and not the whole generation history $\mathcal{G}^{(0)} \ldots
\mathcal{G}^{(t-1)}$).
To evaluate it further, we consider the set of generation states $\mathcal{S}$
of all valid state pairs $s = (\mathcal{G}^{(t)}, \node)$ of a partial graph
$\mathcal{G}^{(t)}$ and a focus node $\node$.
\begin{wrapfigure}[10]{r}{0.22\textwidth}
\centering
\vspace{-0.2cm}
\includegraphics[width=0.18\textwidth]{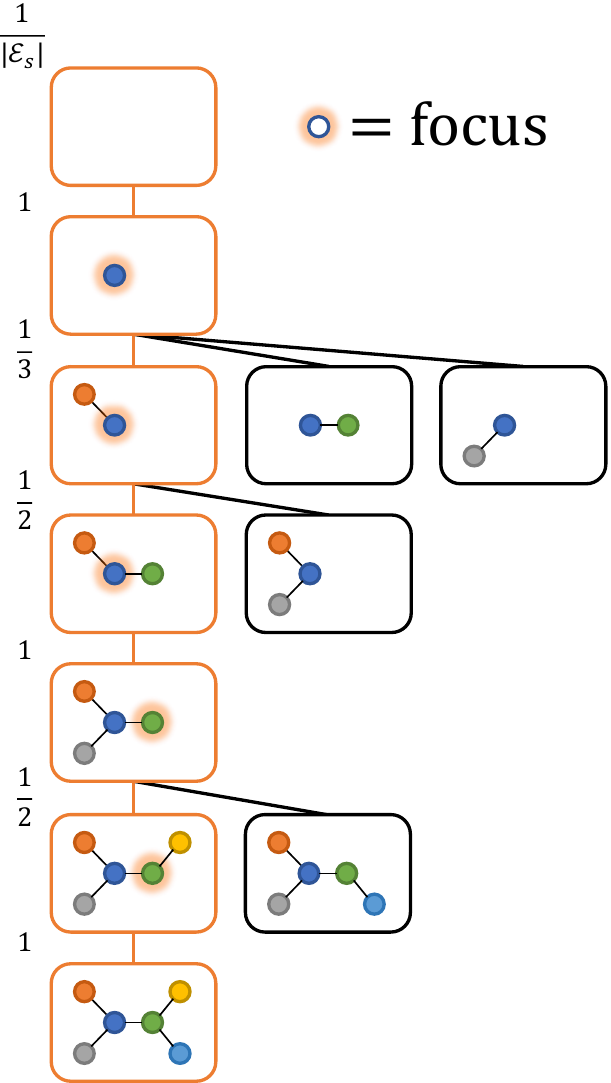}
\caption{Steps considered in our model.}
\label{fig:intuitLoss}
\end{wrapfigure}
We use $|s|$ to denote the multiplicity of state $s$ in $\Pi$, i.e., the number
of traces that contain graph $\mathcal{G}^{(t)}$ and focus on node $v$.
Let $\mathcal{E}_s$ denote all edges that could be generated at state $s$,
i.e., the edges from the focus node $v$ that are present in
the graph $\mathcal{G}$ from the dataset, but are not yet present in
$\mathcal{G}^{(t)}$.
Then, each of these appears uniformly as the next edge to generate in a trace
for all $|s|$ occurrences of $s$ in a trace from $\Pi$,

and therefore, we can rearrange a sum over paths into a sum over steps:
\begin{align*}
\frac{1}{|\Pi|}\sum_{\pi\in \Pi}
  \sum_{(t, \node,\epsilon)\in \pi} 
    \log p(\epsilon \mid s)
&= \frac{1}{|\Pi|}\sum_{s \in \mathcal{S}}
      \sum_{\epsilon \in \mathcal{E}_s} \frac{|s|}{|\mathcal{E}_s|}
      \log p(\epsilon \mid s)\\
&= \mathbb{E}_{s\sim \Pi}\left[\frac{1}{|\mathcal{E}_s|}\sum_{\epsilon \in \mathcal{E}_s}
      \log p(\epsilon \mid s)\right]
\end{align*}
Here we use that $|s|/|\Pi|$ is the probability of observing state $s$ in a
random draw from all states in $\Pi$. We use this expression in
\eqnref{graphLogProb} and train our VAE with a reconstruction loss
$\mathcal{L}_{\rm recon.} =
\sum_{\mathcal{G}\in\mathcal{D}}\log\left[p(\mathcal{G} \mid \mathcal{G}^{(0)})
                                    \cdot p(\mathcal{G}^{(0)} \mid \latent{})\right]$
ignoring additive constants.

We evaluate the expectation over states $s$ using a Monte Carlo estimate from a
set of enumerated generation traces.
In practice, this set of paths is very small (e.g. a single trace) resulting in
a high variance estimate.
Intuitively, \figref{intuitLoss} shows that rather than requiring the model to
exactly reproduce each step of the sampled paths (orange) our objective does not
penalize the model for choosing any valid expansion at each step (black).

\subsection{Optimizing Graph Properties}
\label{sec:optimizing}
So far, we have described a generative model for graphs. In addition, we may wish to perform (local) optimization of these graphs with respect to some numerical property, $Q$. This is achieved by gradient ascent in the continuous latent space using a differentiable gated regression model
\begin{equation*}
	R(\latent{v}) = \sum_v \sigma(g_1(\latent{v})) \cdot g_2(\latent{v}),
\end{equation*}
where $g_1$ and $g_2$ are neural networks\footnote{In our experiments, both $g_1$ and $g_2$ are implemented as linear transformations that project to scalars.} and $\sigma$ is the
sigmoid function. Note that the combination of $R$ with $G_{\rm enc}$ (i.e.,
$R(G_{\rm enc}(\mathcal{G}))$) is exactly the GGNN regression model from
\citet{Gilmer17}.
During training, we use an $L_2$ distance loss $\mathcal{L}_{Q}$ between
$R(\latent{v})$ and the labeled properties $Q$.
This regression objective shapes the latent space, allowing us to optimize for
the property $Q$ in it.
Thus, at test time, we can sample an initial latent point $\latent{v}$ and then
use gradient ascent to a locally optimal point $\latent{v}^*$ subject to an
$L_2$ penalty that keeps the $\latent{v}^*$ within the standard normal prior of
the VAE.
Decoding from the point $\latent{v}^*$ then produces graphs with an optimized
property $Q$.
We show this in our experiments in \secref{qedoptexp}.

\subsection{Training objective}

The overall objective is $\mathcal{L} = \mathcal{L}_{\rm recon.} + \lambda_1 \mathcal{L}_{\rm latent} + \lambda_2 \mathcal{L}_{Q}$, consisting of the usual VAE objective (reconstruction terms and regularization on the latent variables) and the regression loss. Note that we allow deviation from the pure VAE loss ($\lambda_1 = 1$) following \citet{yeung2017tackling}.

\section{Application: Molecule Generation}
In this section, we describe additional specialization of our model for the application of generating chemical molecules. Specifically, we outline details of the molecular datasets that we use and the domain specific masking factors that appear in \eqnref{factoredProb}.

\subsection{Datasets}
\label{sec:datasets}
We consider three datasets commonly used in the evaluation of computational
chemistry approaches:
\begin{itemize}
  \item QM9~\citep{ruddigkeit2012enumeration, ramakrishnan2014quantum}, an enumeration of $\sim{}134$k stable organic
    molecules with up to 9 heavy atoms (carbon, oxygen, nitrogen and fluorine).
    As no filtering is applied, the molecules in this dataset only reflect basic
    structural constraints.
  \item ZINC dataset~\citep{irwin2012zinc}, a curated set of 250k commercially available drug-like chemical
    compounds.
    On average, these molecules are bigger ($\sim{}23$ heavy atoms)
    and structurally more complex than the molecules in QM9.
  \item CEPDB~\citep{hachmann2011harvard, cepdb}, a dataset of organic molecules with an
    emphasis on photo-voltaic applications.
    The contained molecules have $\sim{}28$ heavy atoms on average and contain six to
    seven rings each. We use a subset of the full database containing 250k randomly sampled molecules.
\end{itemize}

For all datasets we kekulize the molecules so that the only edge types to consider are single, double and triple covalent bonds and we remove all hydrogen atoms. In the encoder, molecular graphs are presented with nodes annotated with onehot vectors $\bm{\tau}^*_{\node}$ indicating their atom type and charge.

\subsection{Valency masking}
\label{sec:masking}
Valency rules impose a strong constraint on constructing syntactically valid molecules\footnote{Note that more complex domain knowledge
  e.g. Bredt's rule~\citep{bredt1902ueber} could also be handled in our model but we do not
  implement this here.}. The valency of an atom indicates the number of bonds that that atom can make in a stable molecule, where edge types ``double'' and ``triple'' count for 2 and 3 bonds respectively. In our data, each node type has a fixed valency given by known chemical properties, for example node type ``O'' (an oxygen atom) has a valency of 2 and node type ``O$^-$'' (an oxygen ion) has valency of 1. Throughout the generation process, we use masks $M$ and $m$ to guarantee that the number of bonds $b_{\node}$ at each node never exceeds the valency $b_{\node}^*$ of the node. If $b_{\node}<b_{\node}^*$ at the end of generation we link $b_{\node}^* - b_{\node}$ hydrogen atoms to node $\node$. In this way, our generation process always produces syntactically valid molecules (we define syntactic validity as the ability to parse the graph to a SMILES string using the RDKit parser \cite{landrum2014rdkit}). More specifically, $\smash{M_{\edge{\node}{\nodeTwo}}^{(t)}}$ also handles avoidance of edge duplication and self loops, and is defined as:
\begin{equation}
  M_{\edge{\node}{\nodeTwo}}^{(t)} =
         \mathbb{1}(b_{\node}<b_{\node}^*)
  \times \mathbb{1}(b_{\nodeTwo}<b_{\nodeTwo}^*)
  \times \mathbb{1}(\text{no } \edge{\node}{\nodeTwo} \text{ exists})
  \times \mathbb{1}(\node \neq \nodeTwo) \times \mathbb{1}(\nodeTwo \mbox{ is not closed}),
\end{equation}
where $\mathbb{1}$ is an indicator function, and as a special case, connections to the stop node are always unmasked. Further, when selecting the label for a chosen edge, we must again avoid violating the valency constraint, so we define $\smash{m_{\labelledEdge{\node}{\nodeTwo}{\ell}}^{(t)} = M_{\edge{\node}{\nodeTwo}}^{(t)} \times \mathbb{1}(b_{\nodeTwo}^* - b_{\nodeTwo} \leq \ell)}$, using $\ell = 1,2,3$ to indicate single, double and triple bond types respectively

 \section{Experiments}
\label{sec:Experiments}
We evaluate baseline models, our model (\ourmodel) and a number of ablations on the two tasks of molecule generation and optimization\footnote{Our implementation of CGVAE can be found at \url{https://github.com/Microsoft/constrained-graph-variational-autoencoder}.}.
  
\subsection{Generating molecules}

\begin{figure}
\centering
\begin{minipage}{.57\textwidth}
	(a)\\ \\
	{\tiny
 \centering
 \setlength{\tabcolsep}{0.5em}
  \begin{tabular}{@{}cl|rrrrrrr@{}}
  \toprule
     &  Measure     & 2: \ourmodel & 3: \cite{li2018learning} & 4: LSTM
     														& 5: \cite{gomez-bombarelli16automatic}
                           & 6: \cite{kusner2017grammar}
                           & 7: \cite{simonovsky2018towards}
                           & 8: \cite{samanta2018designing}
    \\
  \midrule
  \parbox[t]{2mm}{\multirow{3}{*}{\rotatebox[origin=c]{90}{QM9}}}
     & \% valid     & 100    & - & 94.78 & 10.00  & 30.00  & 61.00  & 98.00 \\
     & \% novel     & 94.35  & - & 82.98 & 90.00  & 95.44  & 85.00  & 100   \\    
     & \% unique    & 98.57  & - & 96.94 & 67.50  &  9.30  & 40.90  & 99.86 \\
  \midrule
  \parbox[t]{2mm}{\multirow{3}{*}{\rotatebox[origin=c]{90}{ZINC}}}
     & \% valid     & 100    & 89.20	& 96.80 & 17.00  & 31.00  & 14.00  & - \\
     & \% novel     & 100    & 89.10	& 100 & 98.00  & 100    & 100    & - \\    
     & \% unique    & 99.82  & 99.41	& 99.97 & 30.98  & 10.76  & 31.60  & - \\
  \midrule
  \parbox[t]{2mm}{\multirow{3}{*}{\rotatebox[origin=c]{90}{CEPDB}}}
     & \% valid     & 100    & - & 99.61 & 8.30   & 0.00    & -      & -      \\
     & \% novel     & 100    & - & 92.43 & 90.05  & -       & -      & -      \\    
     & \% unique    & 99.62  & - & 99.56 & 80.99  & -       & -      & -      \\
  \bottomrule 
  \end{tabular}
  }\\ \vspace{0.1cm} \\
  (c)
  \begin{center}
  \includegraphics[width=0.9\textwidth]{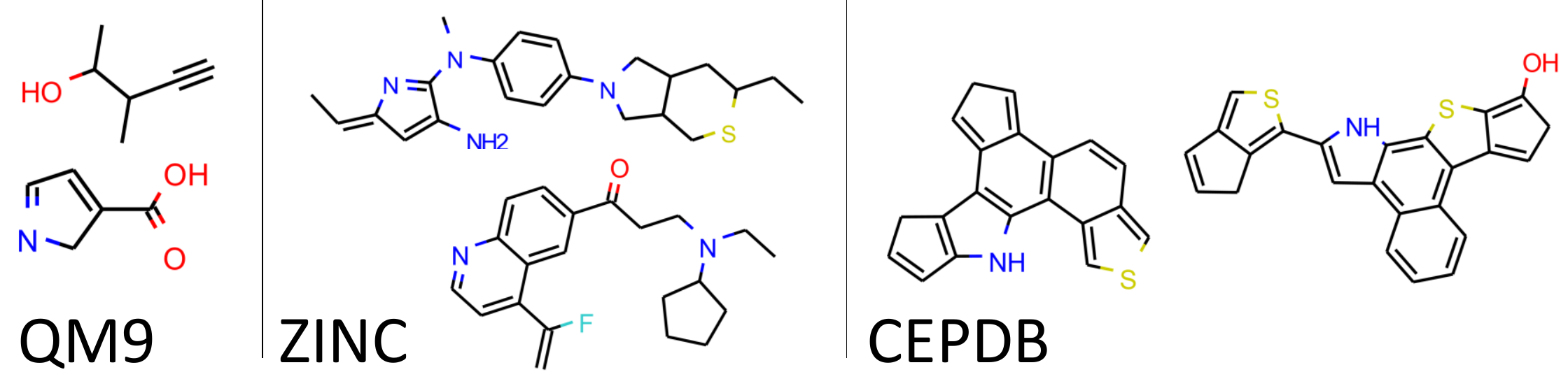}
  \end{center}
\end{minipage}
\begin{minipage}{.42\textwidth}
	(b)\\ 
	\begin{center}
	\includegraphics[width=0.9\textwidth]{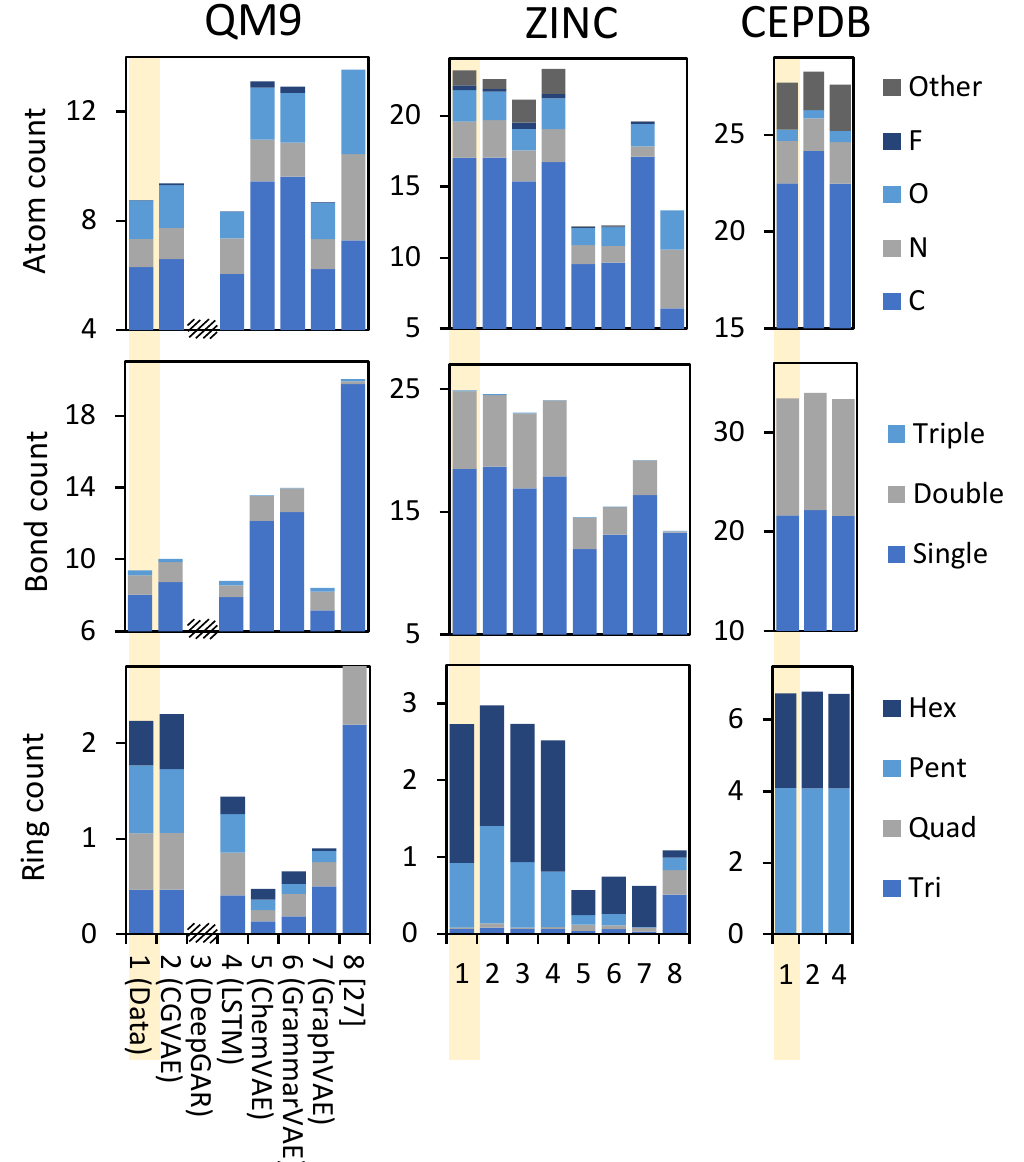}
	\end{center}
\end{minipage}
\caption{Overview of statistics of sampled
molecules from a range of generative models trained on different datasets. In (b) We highlight the target statistics of the datasets in yellow and use the numbers $2,...,7$ to denote different models as shown in the axis key. A hatched box indicates where other works do not supply benchmark results. Two samples from our model on each dataset are shown in (c), with more random samples given in supplementary material \ref{app:samples}.
\label{fig:ResultsBaselines}}
\end{figure}

As baseline models, we consider
 the deep autoregressive graph model (that we refer to as \yujia) from \cite{li2018learning},
 a SMILES generating LSTM language model with 256 hidden units (reduced to 64 units for the smaller QM9 dataset),
 ChemVAE~\cite{gomez-bombarelli16automatic},
 GrammarVAE~\cite{kusner2017grammar},
 GraphVAE~\cite{simonovsky2018towards},
 and the graph model from \cite{samanta2018designing}.
We train these and on our three datasets
and then sample 20k molecules from the trained models (in the case of
\cite{samanta2018designing, li2018learning}, we obtained sets of sampled molecules from the authors). 

We analyze the methods using two sets of metrics. First in \figref{ResultsBaselines}(a) we show metrics from existing work: syntactic validity, novelty (i.e. fraction of sampled molecules not appearing in the training data) and uniqueness (i.e. ratio of sample set size before and after deduplication of identical molecules). Second, in \figref{ResultsBaselines}(b) we introduce new metrics to measure how well each model captures the distribution of molecules in the training set. Specifically, we measure the average number of each atom type and each bond type in the sampled molecules, and we count the average number of 3-, 4-, 5-, and 6-membered cycles in each molecule. This latter metric is chemically relevant because 3- and 4-membered rings are typically unstable due to their high ring strain. \figref{ResultsBaselines}(c) shows 2 samples from our model for each dataset and we show more samples of generated molecules in the supplementary material.

The results in \figref{ResultsBaselines} show that
\ourmodel is excellent at matching graph statistics,
while generating valid, novel and unique molecules for all datasets considered (additional details are found in supplementary material \ref{app:multiplePaths} and \ref{app:additionalProperties}).
The only competitive baselines are \yujia from \citet{li2018learning} and an LSTM language model. Our approach
has three advantages over these baselines: First, whereas >10\% of ZINC-like molecules generated
by \yujia are invalid, our masking
mechanism guarantees molecule validity. An LSTM is surprisingly effective at generating valid molecules, however, 
LSTMs do not permit the injection of domain knowledge (e.g. valence rules or requirement for the existance of 
a particular scaffold) because meaningful constraints cannot be imposed on the flat SMILES representation during generation.
 Second, we train
a shallow model on breadth-first steps rather than full paths and
therefore do not experience 
problems with training instability or overfitting 
that are described in \citet{li2018learning}. Empirical indication for overfitting
in \yujia
is seen by the fact that \citet{li2018learning}
achieves the lowest novelty
score on the ZINC dataset, suggesting that it more
often replays memorized construction traces. It is also observed in the LSTM case, 
where on average 60\% of each generated SMILES string is copied from the nearest neighbour in the training set. 
Converting our generated graphs to SMILES strings reveals only 40\% similarity to the nearest neighbour in the same metric.
Third we are able to use
our continuous latent space for molecule optimization (see below).

\begin{wrapfigure}[13]{r}{0.3\textwidth}
\centering
\vspace{-0.2cm}
\includegraphics[width=0.3\textwidth]{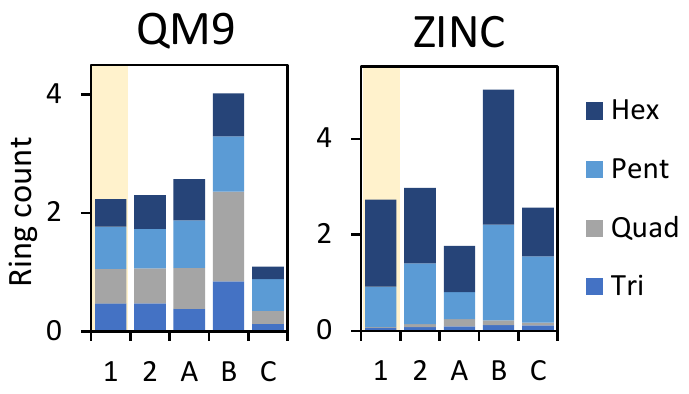}
\caption{Ablation study using the ring metric. 1 indicates statistics of the datasets, 2 of our model and A,B,C of the ablations discussed in the text.}
\label{fig:ablation}
\end{wrapfigure}

We also perform an ablation study on our method. For brevity we only report results
using our ring count metrics, and other
statistics show similar behavior. From all our experiments we highlight three aspects that are
important choices to obtain good results, and we report these in
ablation experiments A, B and C in \figref{ablation}. In experiment A we remove
the distance feature $d_{\node, \nodeTwo}$ from $\bm{\phi}$ and see that this harms performance
on the larger molecules in the ZINC dataset. More interestingly, we see poor results in
experiment B where we make an independence assumption on edge generation (i.e. use features $\bm{\phi}$
to calculate independent probabilities for all possible edges and sample an entire
molecule in one step). We also see poor results in experiment C where we remove the GGNN from the decoder (i.e. perform sequential
construction with $\smash{\nodestate{t}{\node}=\nodestate{0}{\node}}$).
This indicates that the choice to perform sequential decoding with GGNN node updates before each
decision are the keys to the success of our model.

\vspace{-0.3cm}
\subsection{Directed molecule generation}
\label{sec:qedoptexp}
\begin{figure}[t]
 \resizebox{\columnwidth}{!}{%
   \begin{tabular}{@{}lcccccc@{}}
       & \includegraphics[width=0.15\textwidth]{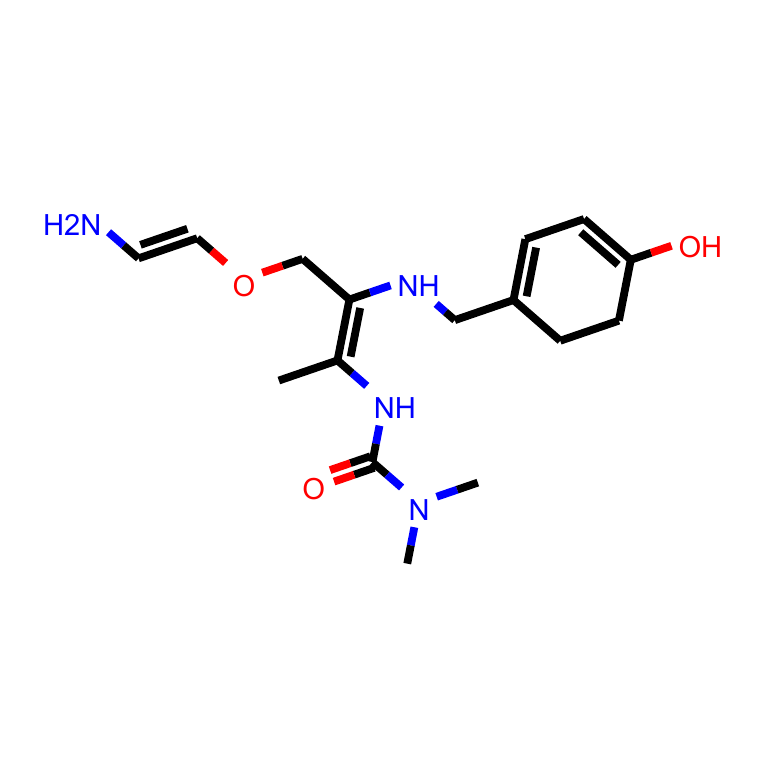}
       & \includegraphics[width=0.15\textwidth]{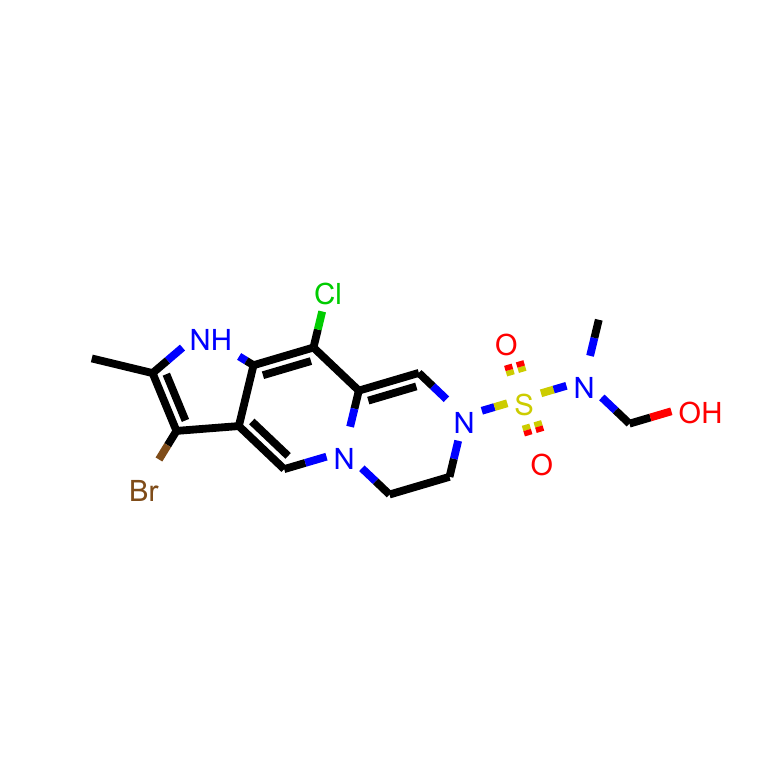}
       & \includegraphics[width=0.15\textwidth]{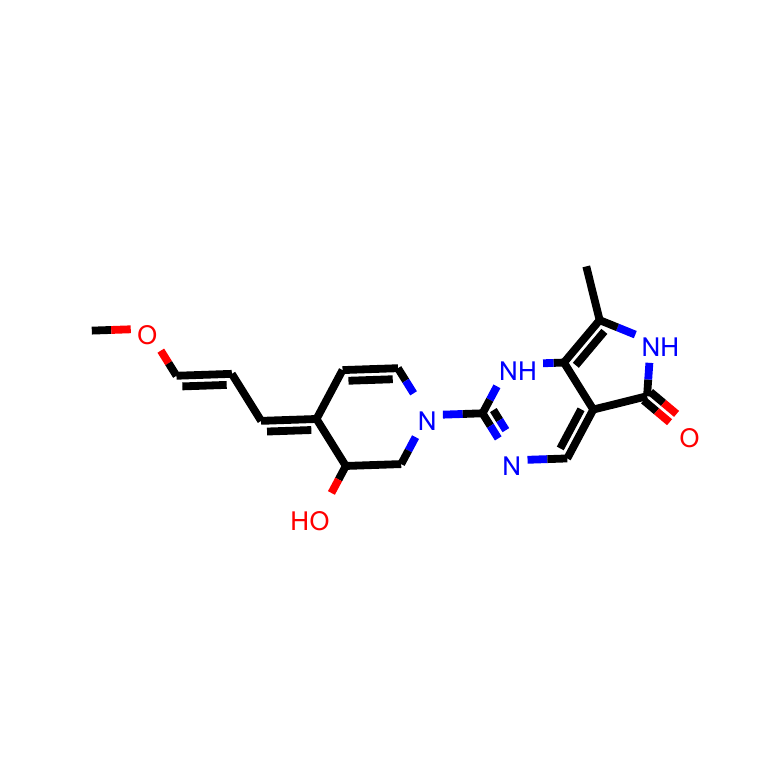}
       & \includegraphics[width=0.15\textwidth]{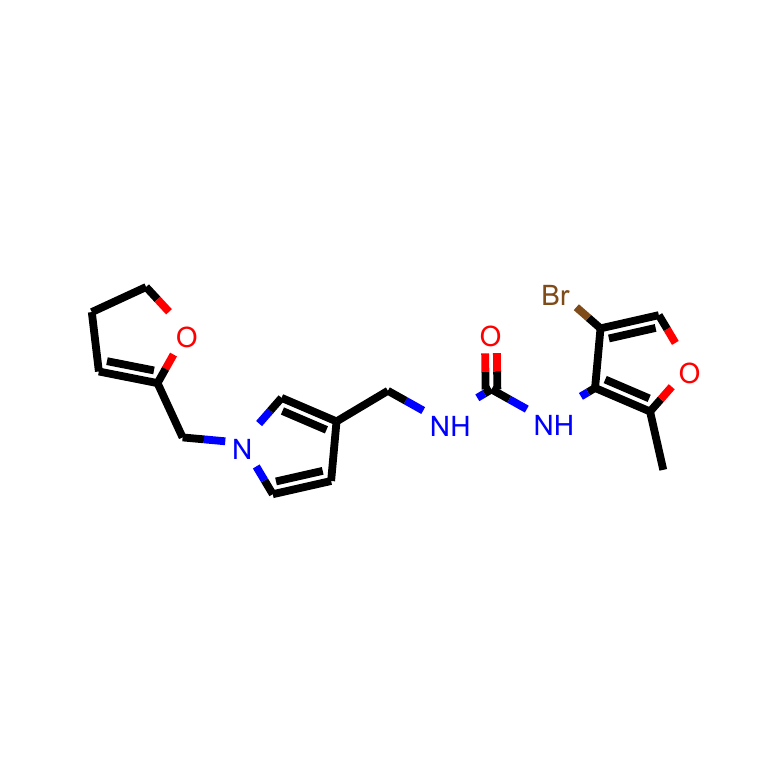}
       & \includegraphics[width=0.15\textwidth]{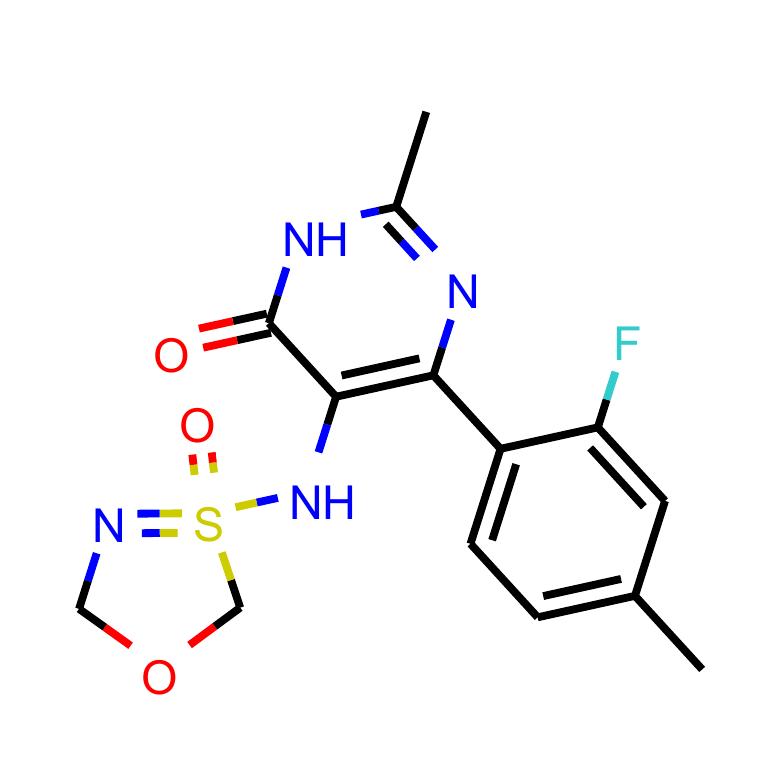}
       & \includegraphics[width=0.15\textwidth]{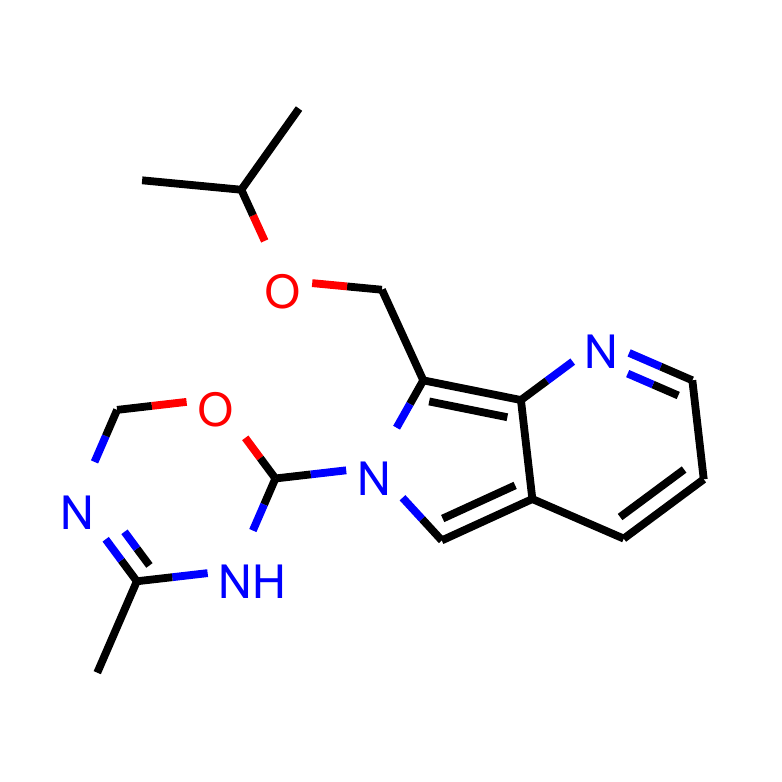}\\
     Pred. QED
       & 0.5686
       & 0.6685
       & 0.7539
       & 0.8376
       & 0.9013
       & 0.9271 \\
     Real QED
       & 0.5345
       & 0.6584
       & 0.7423
       & 0.8298
       & 0.8936
       & 0.9383
   \end{tabular}  
 }
 \caption{\label{fig:opt sequence}Trajectory of QED-directed optimization in
   latent space. Additional examples are shown in supplementary material \ref{app:trajectories}.}
\end{figure}

Finally, we show that we can use the VAE structure of our method to direct the
molecule generation towards especially interesting molecules.
As discussed in \secref{optimizing} (and first shown by \citet{gomez-bombarelli16automatic} in this setting), we extend our
architecture to predict the \emph{Quantitative Estimate of Drug-Likeness} (QED)
directly from latent space.
This allows us to generate molecules with very high QED values by performing
gradient ascent in the latent space using the trained QED-scoring network.
\figref{opt sequence} shows an interpolation sequence from a point in latent
space with an low QED value (which ranges between 0 and 1) to the local
maximum.
For each point in the sequence, the figure shows a generated molecule, the QED
value our architecture predicts for this molecule, as well as the QED value
computed by RDKit.

\section{Conclusion}
We proposed \ourmodel, a sequential generative model for graphs built from a VAE with GGNNs in the encoder and decoder. Using masks that enforce chemical rules, we specialized our model to the application of molecule generation and achieved state-of-the-art generation and optimization results. We introduced basic statistics to validate the quality of our generated molecules. Future work will need to link to the chemistry community to define additional metrics that further guide the construction of models and datasets for real-world molecule design tasks.

\bibliographystyle{abbrvnat}
\bibliography{references}

\newpage
\appendix
{\Large{Supplementary Material: CGVAE for Molecule Design}}
\section{Molecule Samples}
\label{app:samples}
We provide 25 random samples from our model for qualitative comparison with samples from each training dataset.

{\centering
\begin{tabular}{@{}c|cc}
	\toprule
		&	Data Samples	&	CGVAE Samples \\
	\midrule
	\parbox[t]{2mm}{\rotatebox[origin=c]{90}{QM9}} &
	\raisebox{-.5\height}{\includegraphics[width=0.45\textwidth]{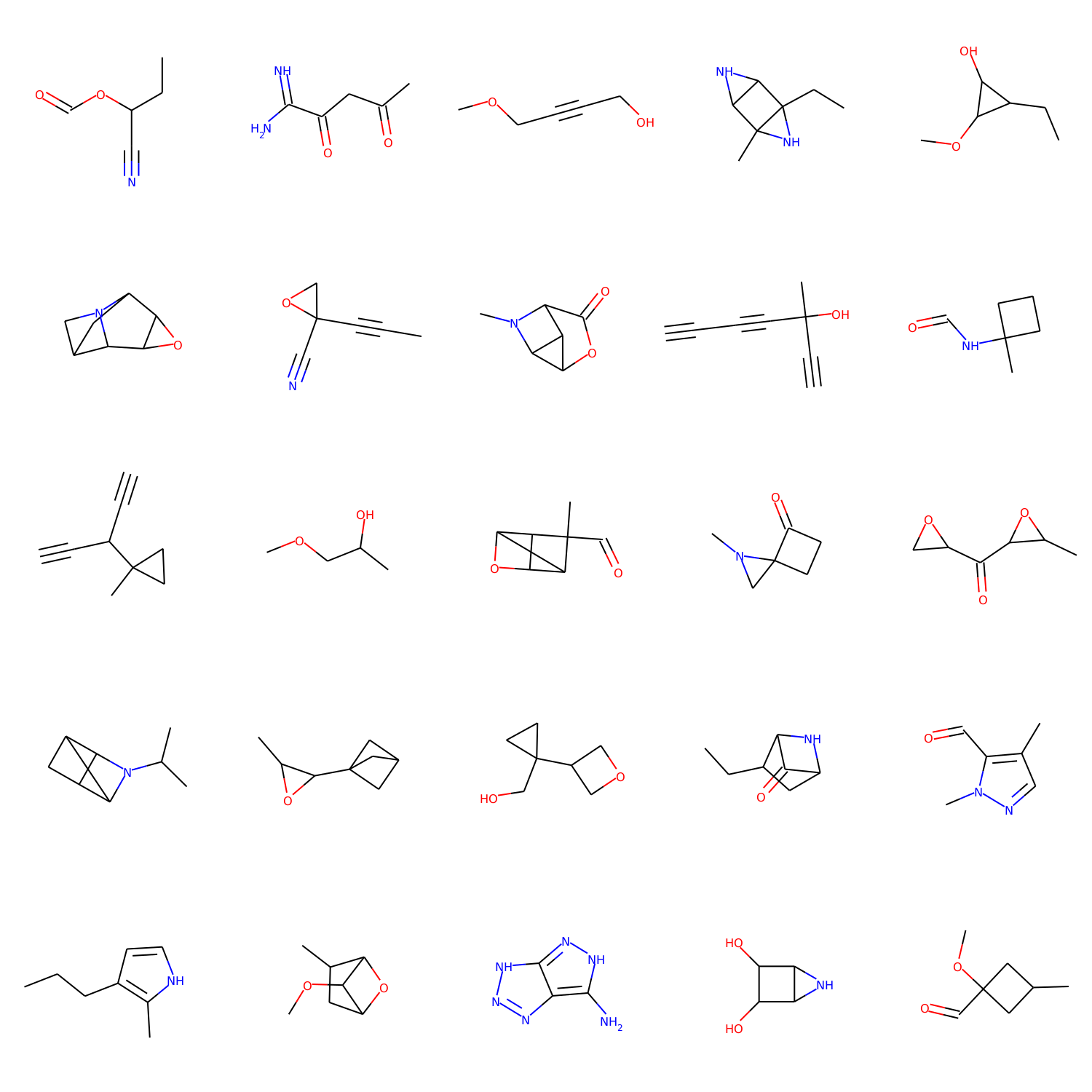}} &
	\raisebox{-.5\height}{\includegraphics[width=0.45\textwidth]{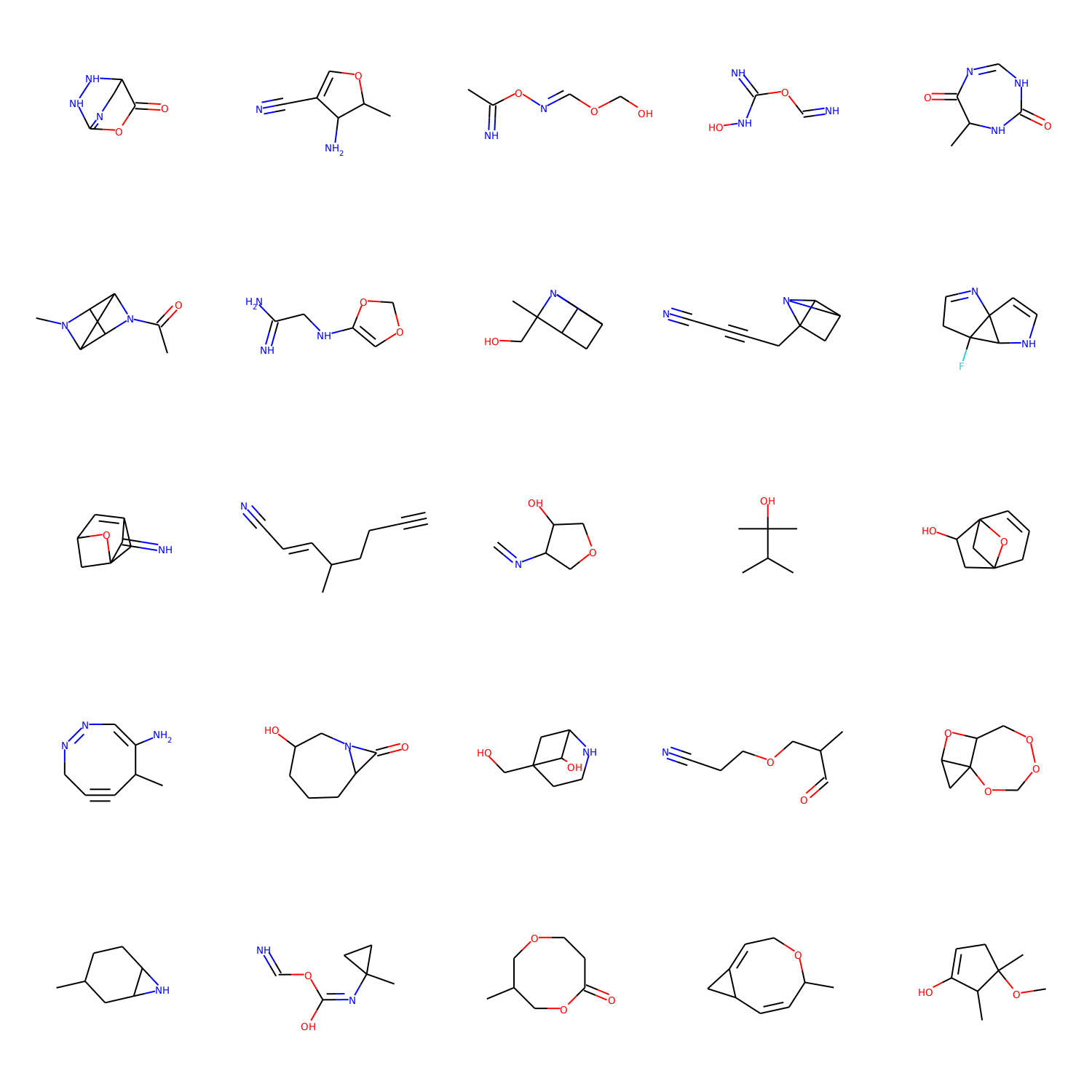}} \\
	\midrule
	\parbox[t]{2mm}{\multirow{1}{*}{\rotatebox[origin=c]{90}{ZINC}}} &
	\raisebox{-.5\height}{\includegraphics[width=0.45\textwidth]{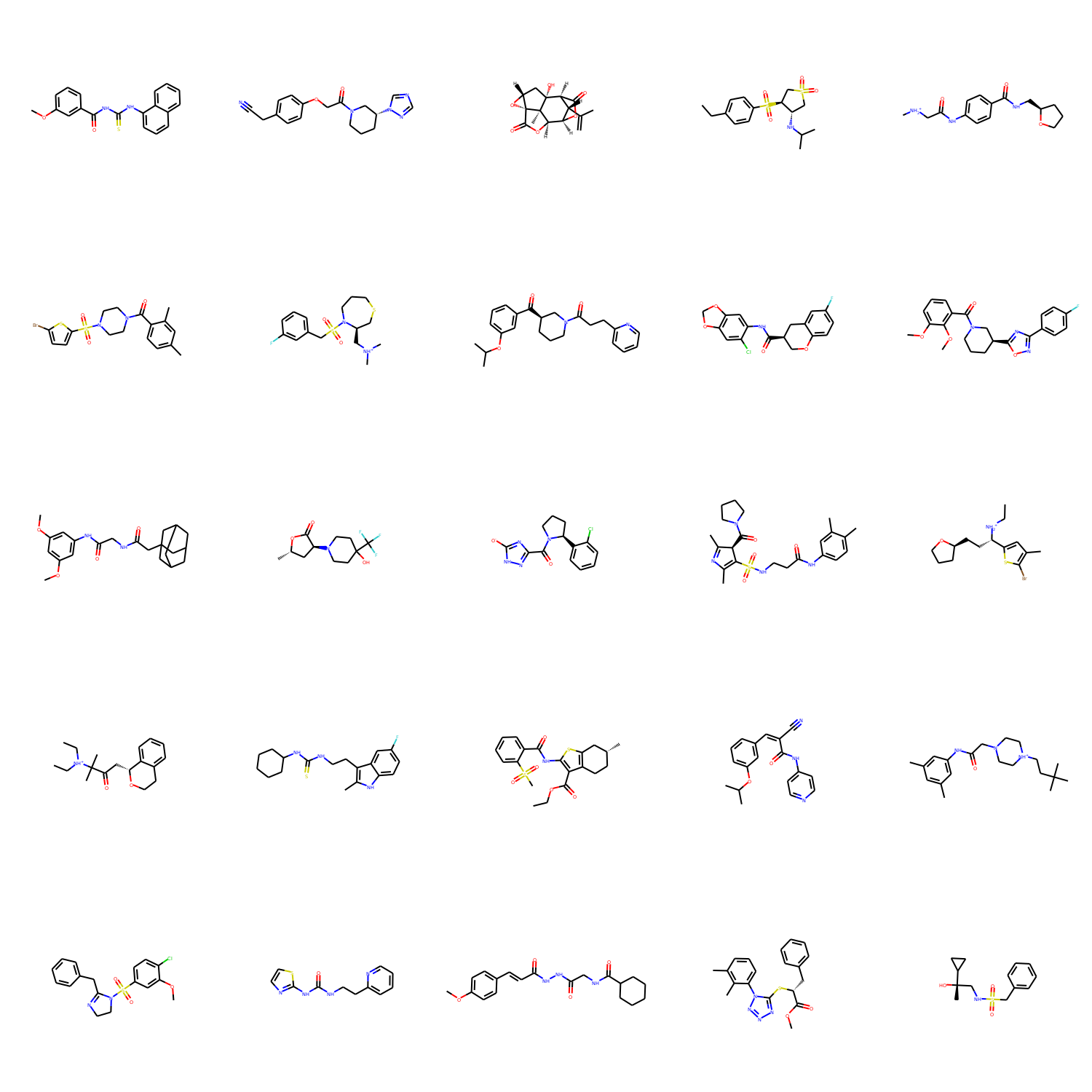}} &
	\raisebox{-.5\height}{\includegraphics[width=0.45\textwidth]{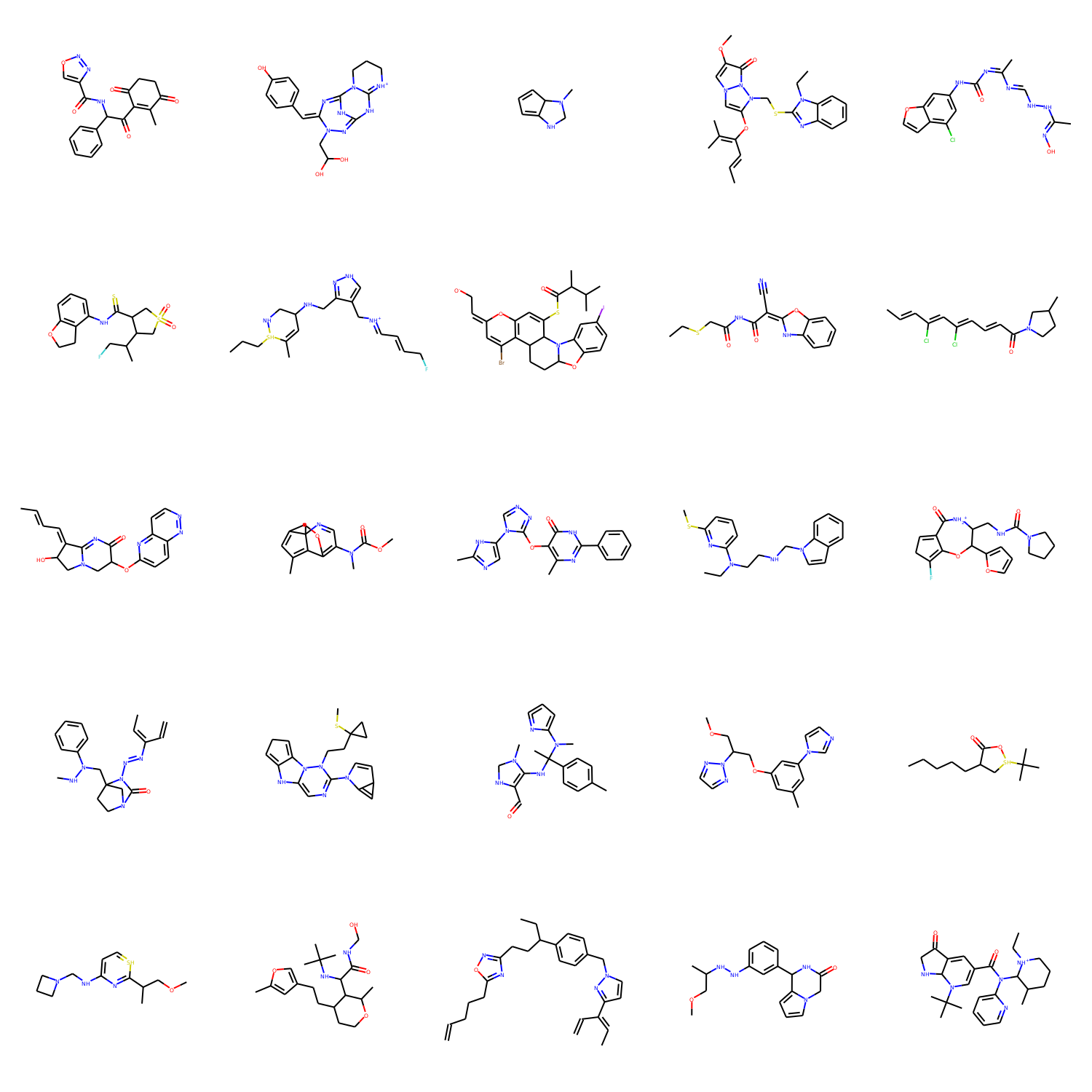}} \\
	\midrule
	\parbox[t]{2mm}{\multirow{1}{*}{\rotatebox[origin=c]{90}{CEPDB}}} &
	\raisebox{-.5\height}{\includegraphics[width=0.45\textwidth]{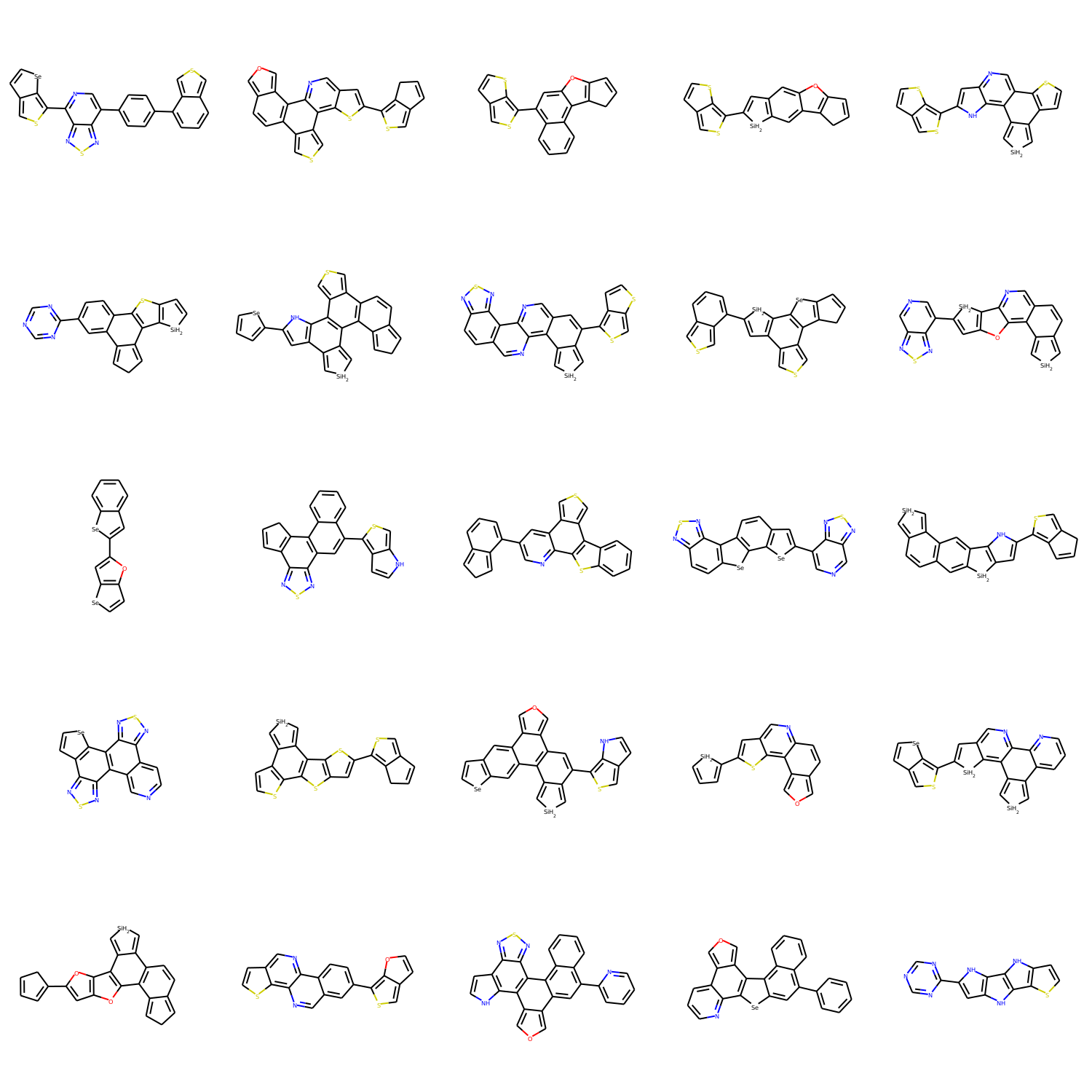}} &
	\raisebox{-.5\height}{\includegraphics[width=0.45\textwidth]{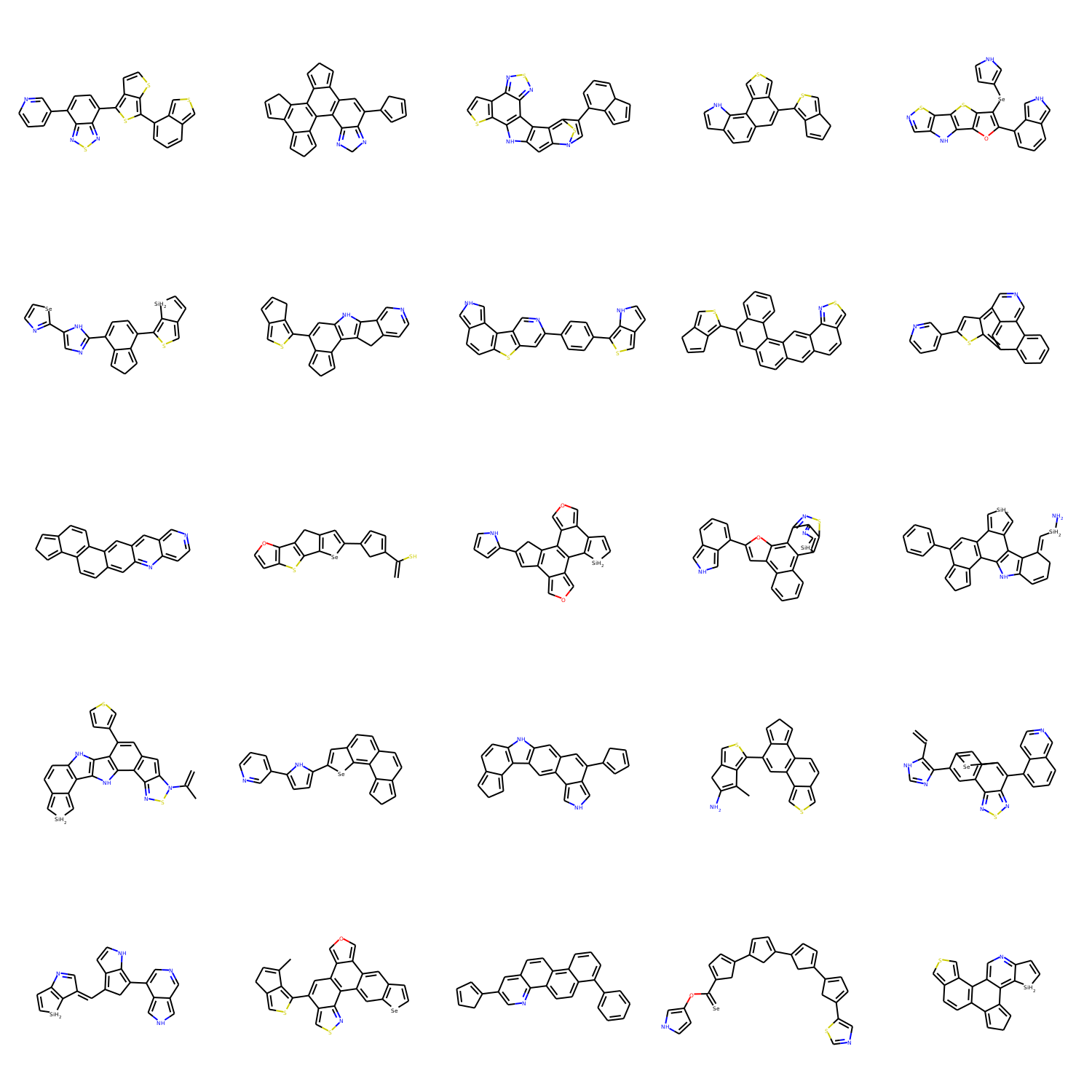}} \\
	\bottomrule
\end{tabular}
}

\section{Effect of multiple training paths}
\label{app:multiplePaths}
\begin{figure}[h!]
	\centering
	\vspace{-0.5cm}
	\includegraphics[width=0.45\textwidth]{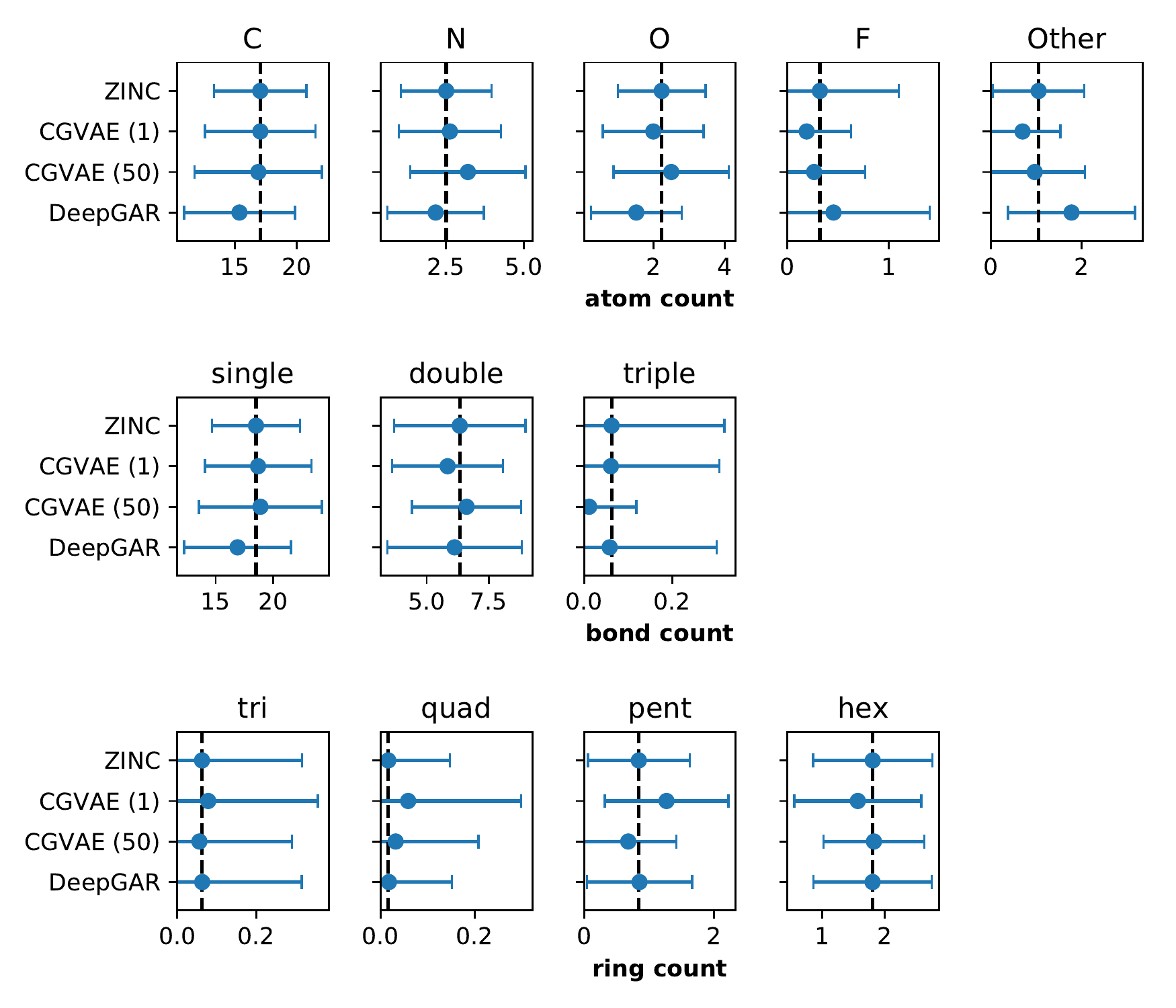}
	\caption{Distribution of structural metrics for models trained on ZINC. We represent the distribution of each property over samples as a point at the mean with error bars covering 1 standard deviation. We calculate the metrics on the raw dataset (ZINC) and samples drawn from CGVAE trained on 1 (CGVAE (1)) or 50 (CGVAE (50)) generation traces.}
	\vspace{-0.5cm}
	\label{fig:errorbars}
\end{figure}

Section \ref{sec:Decoder} describes how we should enumerate all breadth first graph generation traces, break these traces into state transitions and then randomly sample state transitions to give the Monte Carlo estimate of the reconstruction loss. However, for computational efficiency, in the presented experiments we provide only a single trace containing $E$ transitions (where $E$ is the number of edges in the final molecule including edges to the stop node). Figure~\ref{fig:errorbars} shows an additional experiment (CGVAE (50)) where we enumerate 50 traces for each molecule and sample $E$ transitions from this enumeration (so the final dataset size is the same). While increasing the number of traces considered produces a small improvement in the matching of ring statistics in the sampled molecules, it is not clear that this benefit is worth the considerable computational overhead required in preparing the dataset.

\newpage

\section{Additional Molecular Properties}
\label{app:additionalProperties}
Here we provide histograms of the following molecular properites of the sampled molecules for our method and the \yujia and LSTM baselines:
\begin{center}
{\tiny
\begin{tabular}{rlp{2in}}
\toprule
Property & RDKit Implementation & Description\\
\midrule
Molecular Weight & \texttt{Chem.Descriptors.MolWt} & The isotope-averaged molecular weight in atomic mass units. \\
Bertz Complexity & \texttt{Chem.GraphDescriptors.BertzCT} & A topological index meant to quantify complexity of molecules.\\
H donor count & \texttt{Chem.Lipinski.NumHDonors} & Number of heavy atoms bonded to H atoms that can form Hydrogen bonds.\\
H acceptor count & \texttt{Chem.Lipinski.NumHAcceptors} & Number of heavy atoms with lone electron pairs that can form Hydrogen bonds.\\
Rotatable bond count & \texttt{Chem.Lipinski.NumRotatableBonds} & Rotatable bonds are any single bond, not in a ring, bound to a nonterminal heavy atom.\\
Partition coefficient & \texttt{Chem.Crippen.MolLogP} & The octanol/water logP partition coefficient according to Wildman and Crippen 1999.\\
Topological polar surface area & \texttt{Chem.rdMolDescriptors.CalcTPSA} & The total exposed surface area of polar atoms in a molecule including attached Hydrogens (in square angstroms).\\
\bottomrule
\end{tabular}
}
\includegraphics[width=\textwidth]{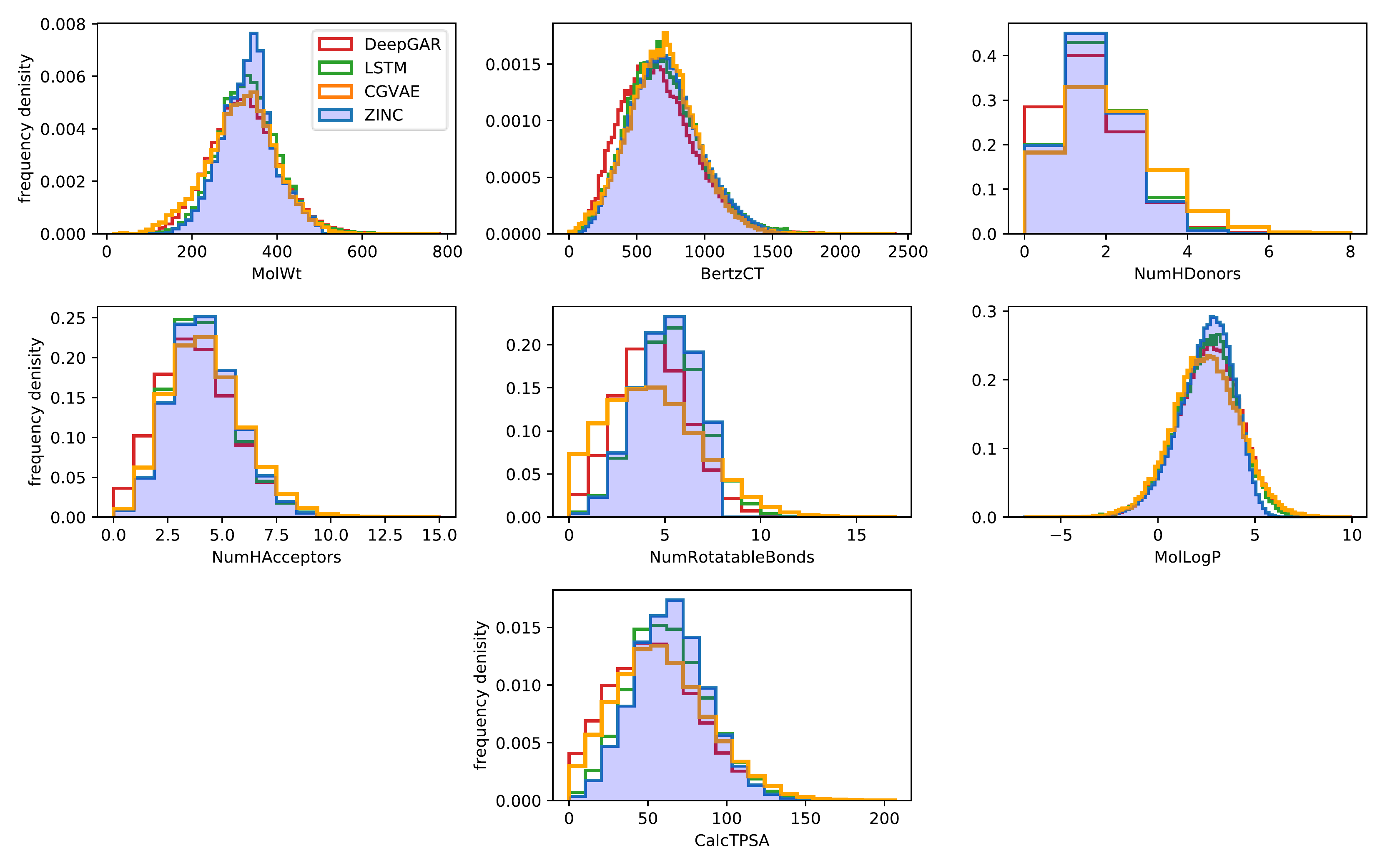}
\end{center}

\section{Optimization trajectories}
\label{app:trajectories}
We provide additional QED optimization trajectories for our model trained on the ZINC dataset.

\resizebox{\columnwidth}{!}{%
   \begin{tabular}{@{}lcccccc@{}}
       & \includegraphics[width=0.15\textwidth]{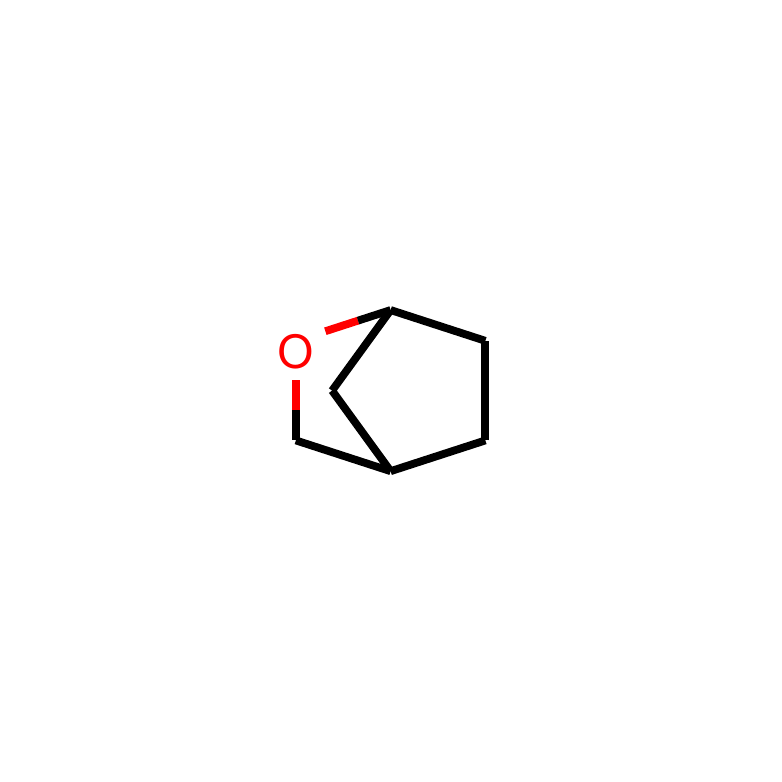}
       & \includegraphics[width=0.15\textwidth]{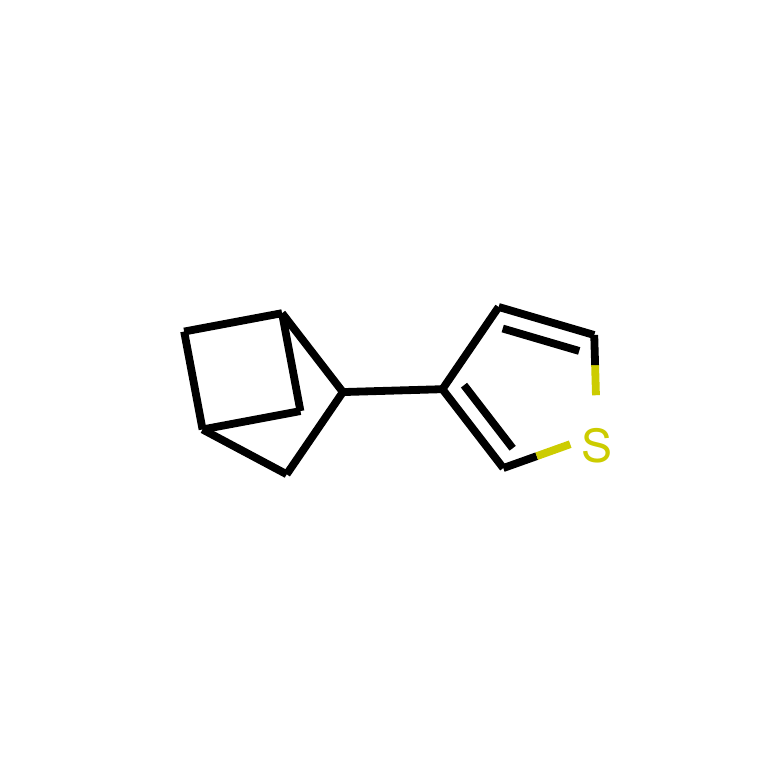}
       & \includegraphics[width=0.15\textwidth]{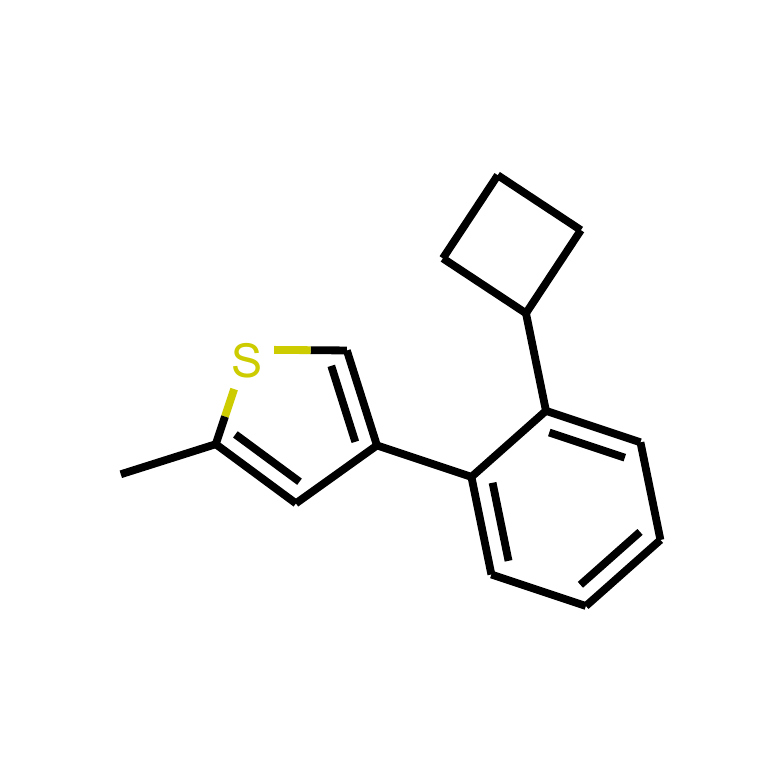}
       & \includegraphics[width=0.15\textwidth]{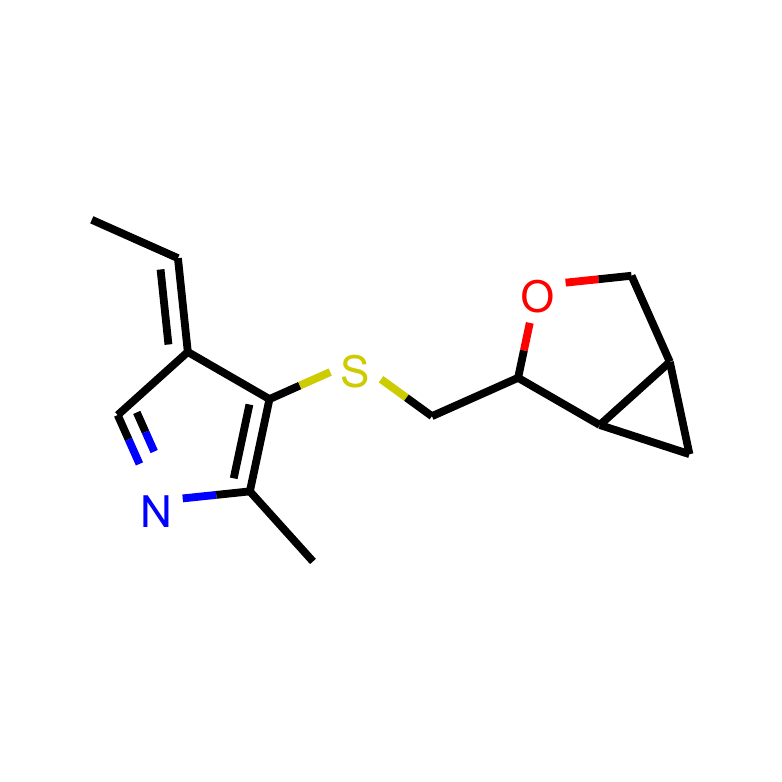}
       & \includegraphics[width=0.15\textwidth]{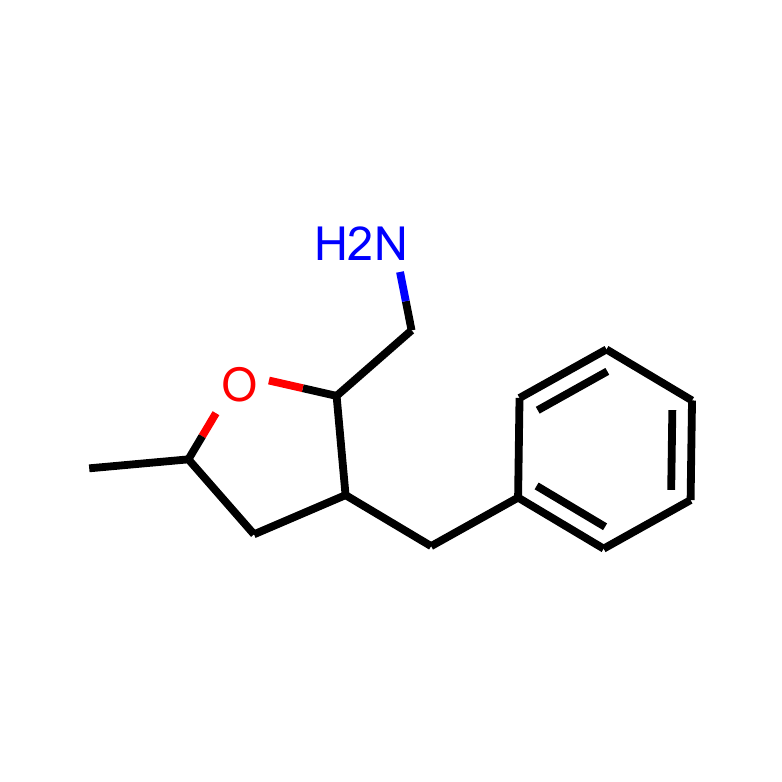}
       & \includegraphics[width=0.15\textwidth]{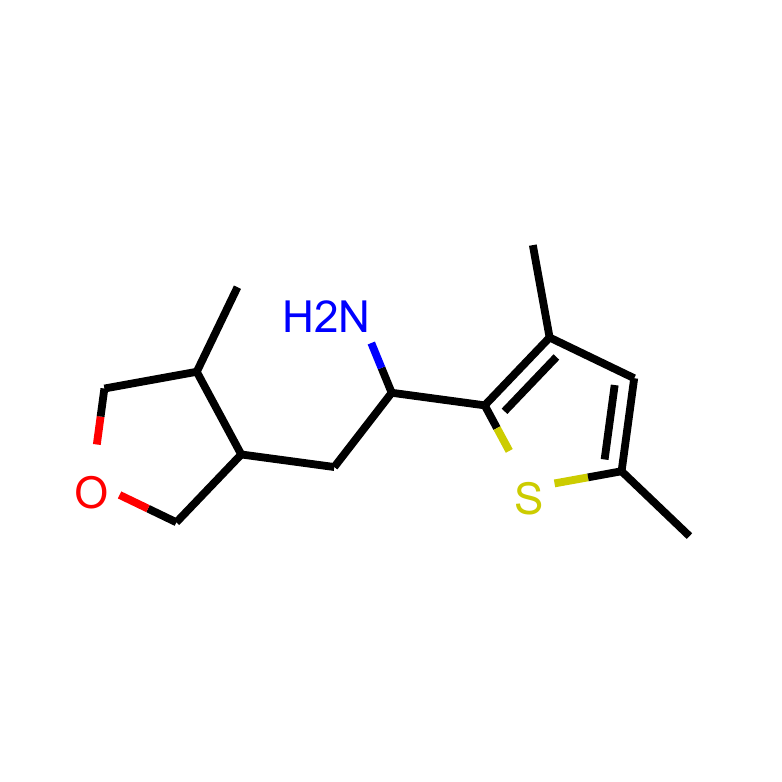}\\
     Pred. QED
       & 0.5073
       & 0.6249
       & 0.6683
       & 0.7375
       & 0.8367
       & 0.8927 \\
     Real QED
       & 0.4419
       & 0.5973
       & 0.6789
       & 0.7497
       & 0.8186
       & 0.8796
   \end{tabular}  
 }
 
 \resizebox{\columnwidth}{!}{%
   \begin{tabular}{@{}lcccccc@{}}
       & \includegraphics[width=0.15\textwidth]{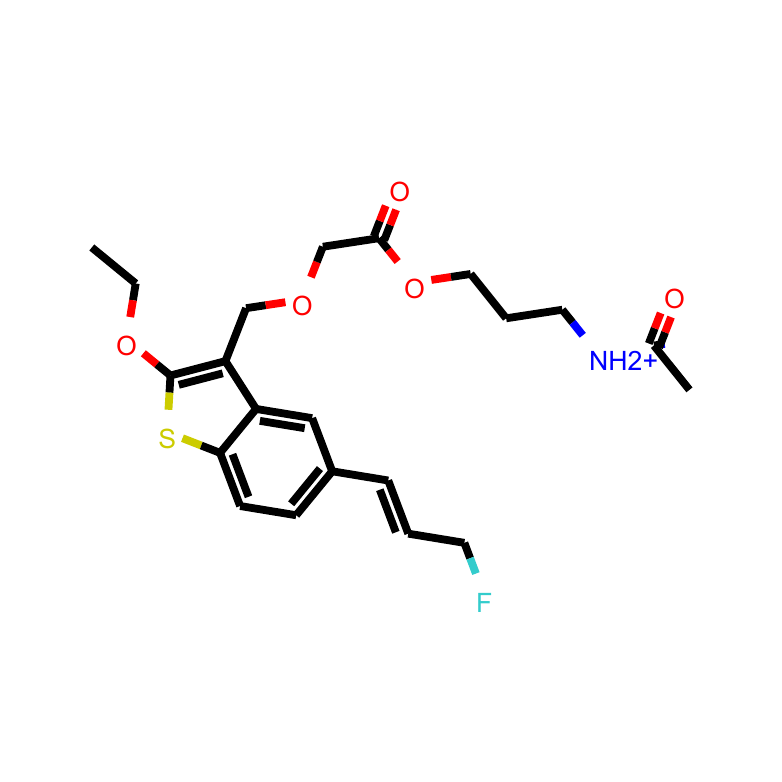}
       & \includegraphics[width=0.15\textwidth]{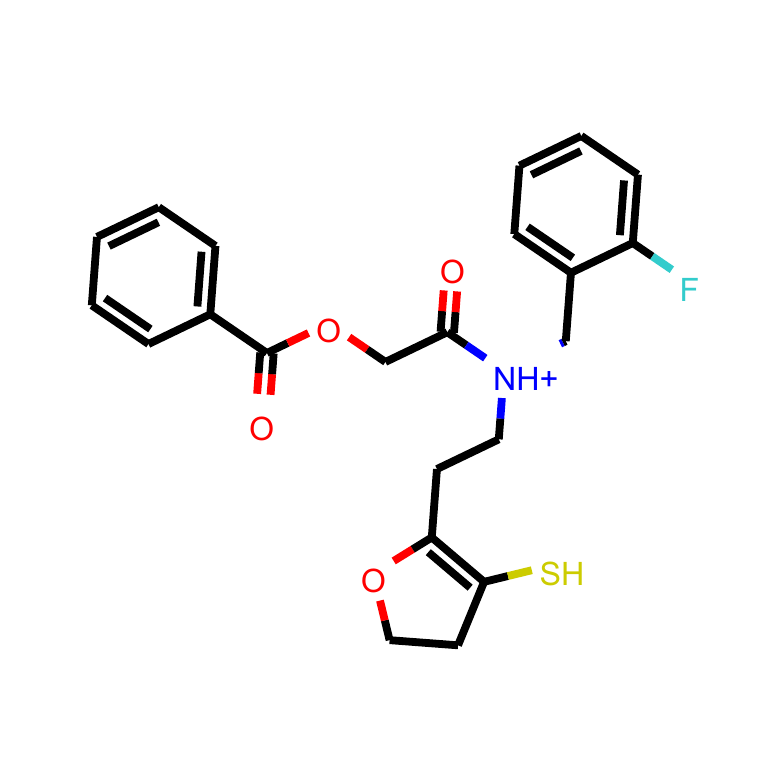}
       & \includegraphics[width=0.15\textwidth]{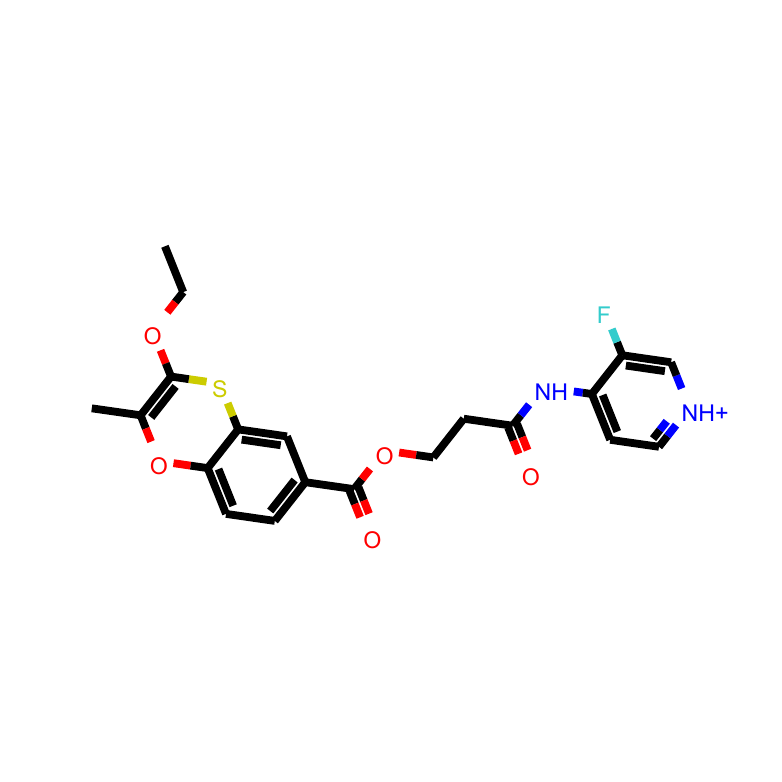}
       & \includegraphics[width=0.15\textwidth]{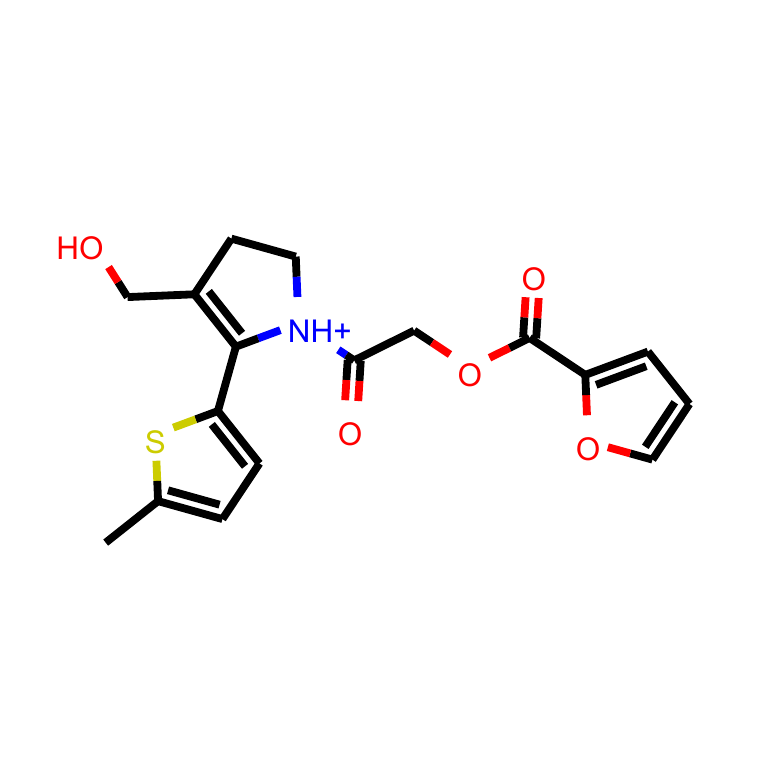}
       & \includegraphics[width=0.15\textwidth]{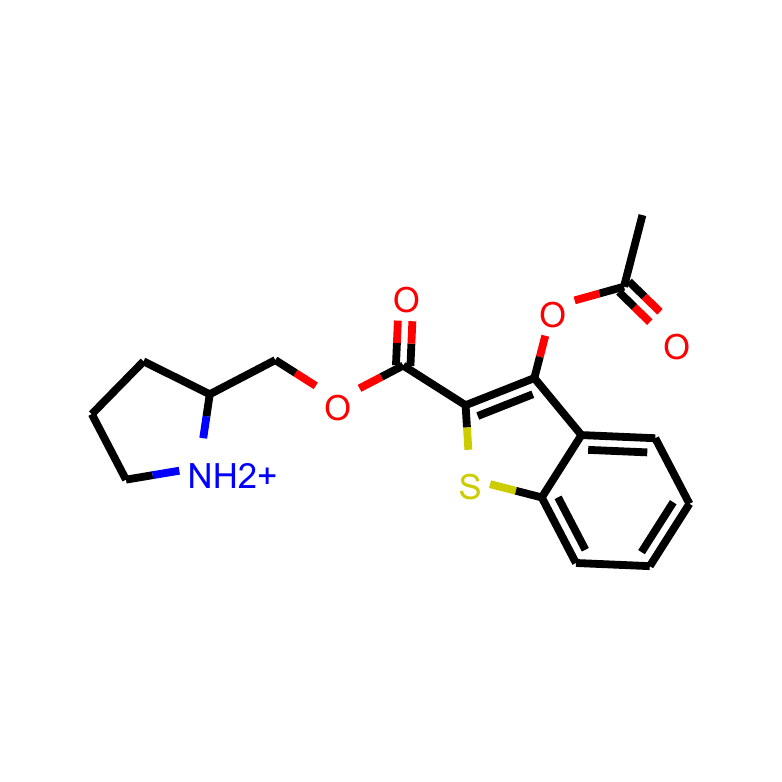}
       & \includegraphics[width=0.15\textwidth]{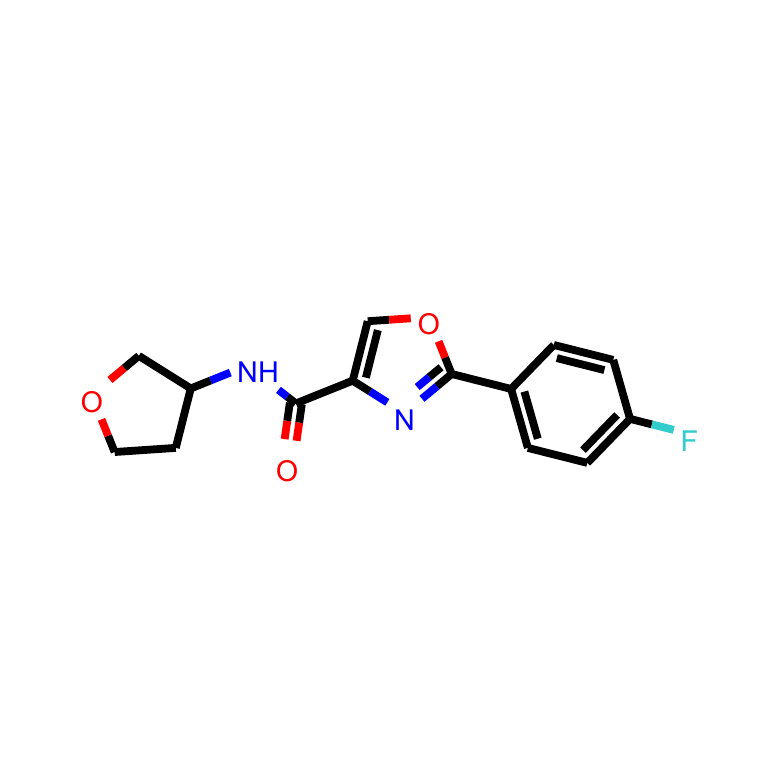}\\
     Pred. QED
       & 0.4368
       & 0.5046
       & 0.7017
       & 0.7683
       & 0.8906
       & 0.9427 \\
     Real QED
       & 0.4188
       & 0.5131
       & 0.6910
       & 0.7889
       & 0.8709
       & 0.9309
   \end{tabular}  
 }

\end{document}